\documentclass[10pt,peerreview,draftclsnofoot,onecolumn,a4paper]{IEEEtran}

\usepackage{amsmath,amssymb}
\usepackage{microtype}
\usepackage{graphicx}
\usepackage[boxed]{algorithm2e}
\usepackage[caption=false,font=footnotesize]{subfig}

\def\i{\boldsymbol{i}}
\def\j{\boldsymbol{j}}
\def\Z{\mathbb{Z}}

\title{A Simple Yet Effective Improvement to the Bilateral Filter for Image Denoising}
\author{Kollipara~Rithwik and Kunal~N.~Chaudhury
\thanks{K.~Rithwik is with the Department of Electrical Engineering, Indian Institute of Technology, Hyderabad, India (e-mail: ee12b1042@iith.ac.in). K.~N.~Chaudhury is with the Department of Electrical Engineering, Indian Institute of Science, Bangalore 560012, India (e-mail: kunal@ee.iisc.ernet.in). R. Kollipara was supported by the Indian Academy of Sciences under the SRPF program, and K.~N.~Chaudhury was partially supported by a Startup Grant from the Indian Institute of Science.}}

\begin{document}

\maketitle

\begin{abstract} 
The bilateral filter has diverse applications in image processing, computer vision, and computational photography. 
In particular, this non-linear filter is quite effective in denoising images corrupted with additive Gaussian noise. 
The filter, however, is known to perform poorly at large noise levels. 
Several adaptations of the filter have been proposed in the literature to address this shortcoming, but often at an added computational cost. 
In this paper, we report a simple yet effective modification that improves the denoising performance of the bilateral filter at almost no additional cost. 
We provide visual and quantitative results on standard test images which show that this improvement is significant both visually and in terms of PSNR and SSIM (often as large as $5$ dB). We also demonstrate how the proposed filtering can be implemented at reduced complexity by adapting a recent idea for fast bilateral filtering.
\end{abstract}


\section{Introduction}

Linear smoothing filters, such as the classical Gaussian filter, typically work well in applications where the amount of smoothing required is small.
For example, they are very quite effective in removing small dosages of additive noise from images. 
However, when the noise floor is large and one is required to average more pixels to improve the signal-to-noise ratio, linear filters tend to
to over-smooth sharp image features such as edges and corners. 
This over-smoothing can be alleviated
using some form of data-driven (non-linear) diffusion, where the quantum of smoothing is controlled using the image features. 
A classical example in this regard is the anisotropic diffusion of Perona and Malik  \cite{Perona1990}. 
The insight of the authors was to take the standard diffusion equation and turn it into a non-linear differential equation by controlling its diffusivity 
using the gradient information. This automatically attenuated the blurring in the vicinity of edges. 
In practice, the associated differential equation is numerically solved using an iterative solver.
While the Perona-Malik diffusion is known to be mathematically ill-posed \cite{Kichenassamy1997}, it is  known to be numerical stable in practice and performs reasonably well on real data. A delicate aspect of this scheme is the choice of the stopping criteria which often critically determines the final result.

\subsection{Bilateral Filter}

\begin{figure*}
\centering
\subfloat[\textit{Barbara}.]{\includegraphics[width=0.25\linewidth]{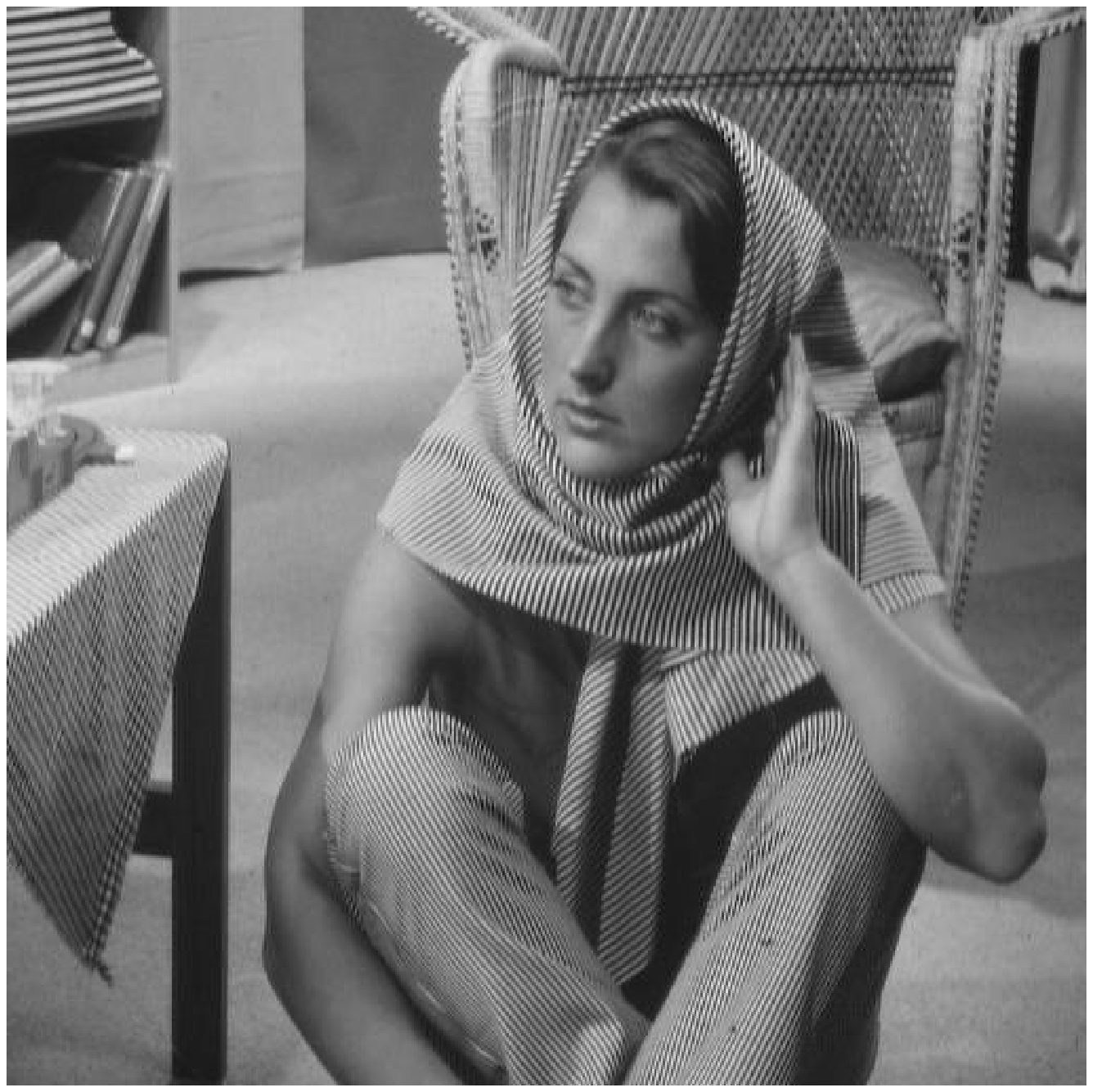}}  
\subfloat[Corrupted ($\sigma=20, 22.11$ dB).]{\includegraphics[width=0.25\linewidth]{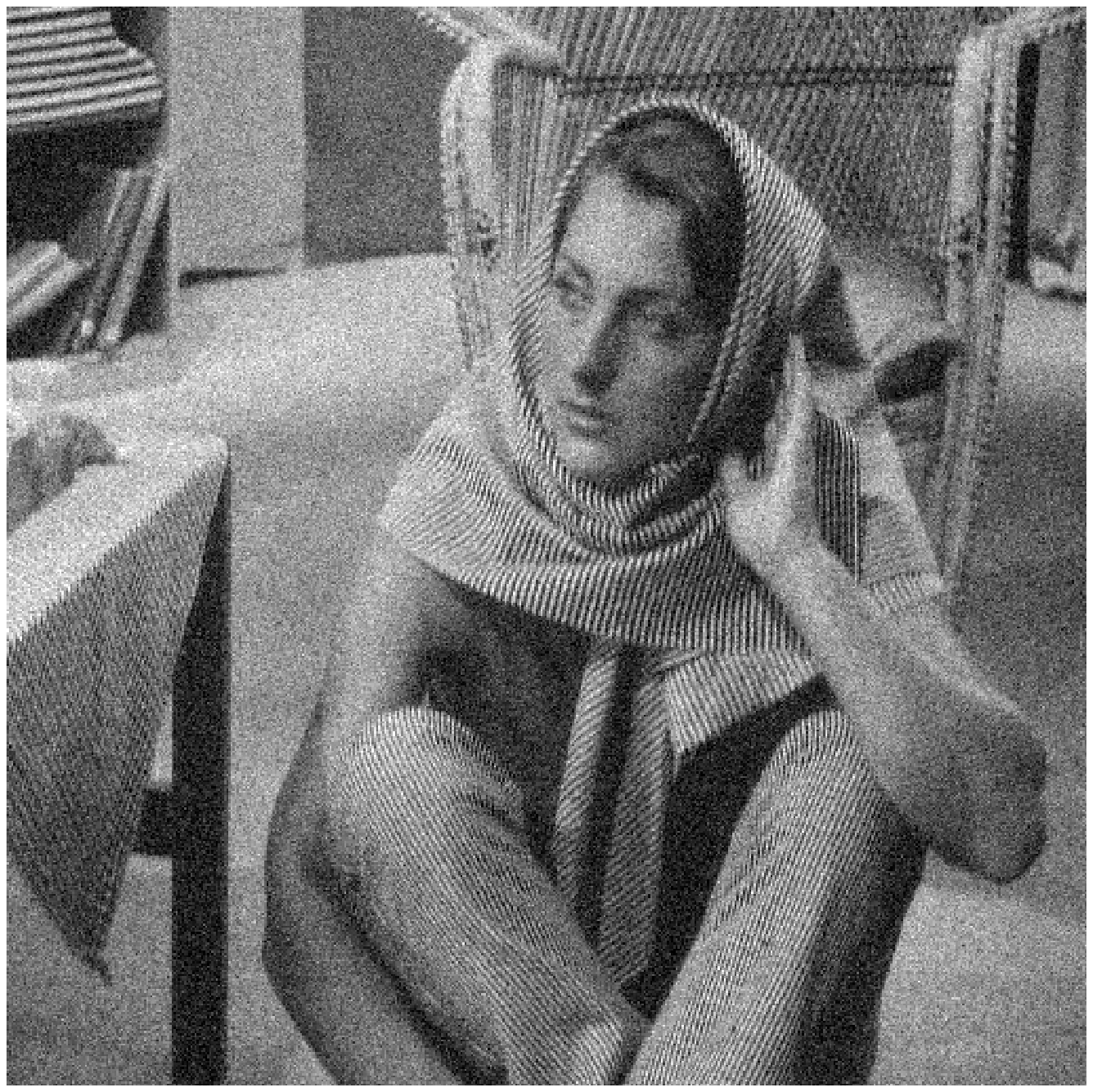}}  
\subfloat[Standard Bilateral ($26.62$ dB).]{\includegraphics[width=0.25\linewidth]{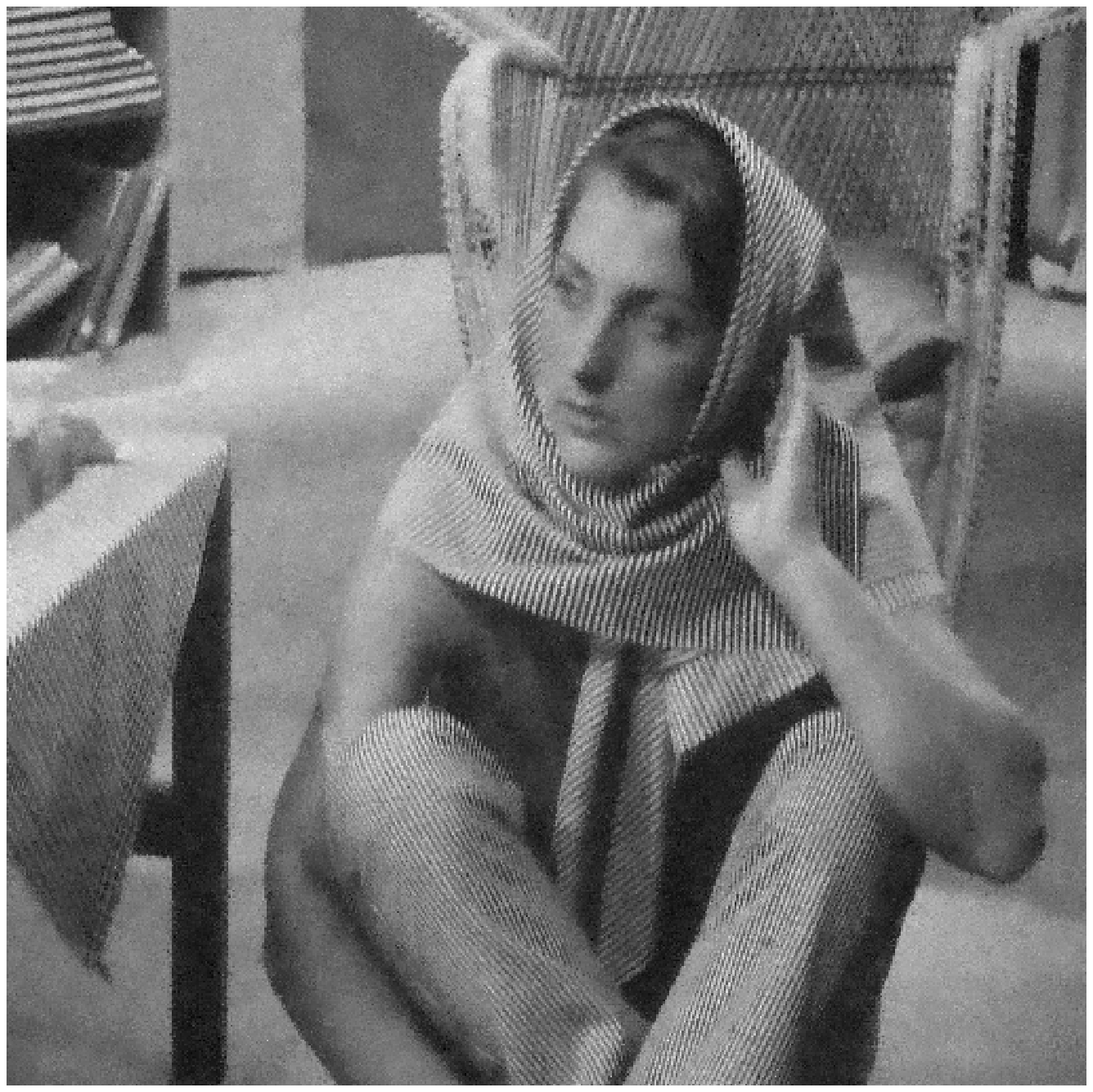}} 
\subfloat[Oracle Bilateral ($\bf{30.05}$ dB).]{\includegraphics[width=0.25\linewidth]{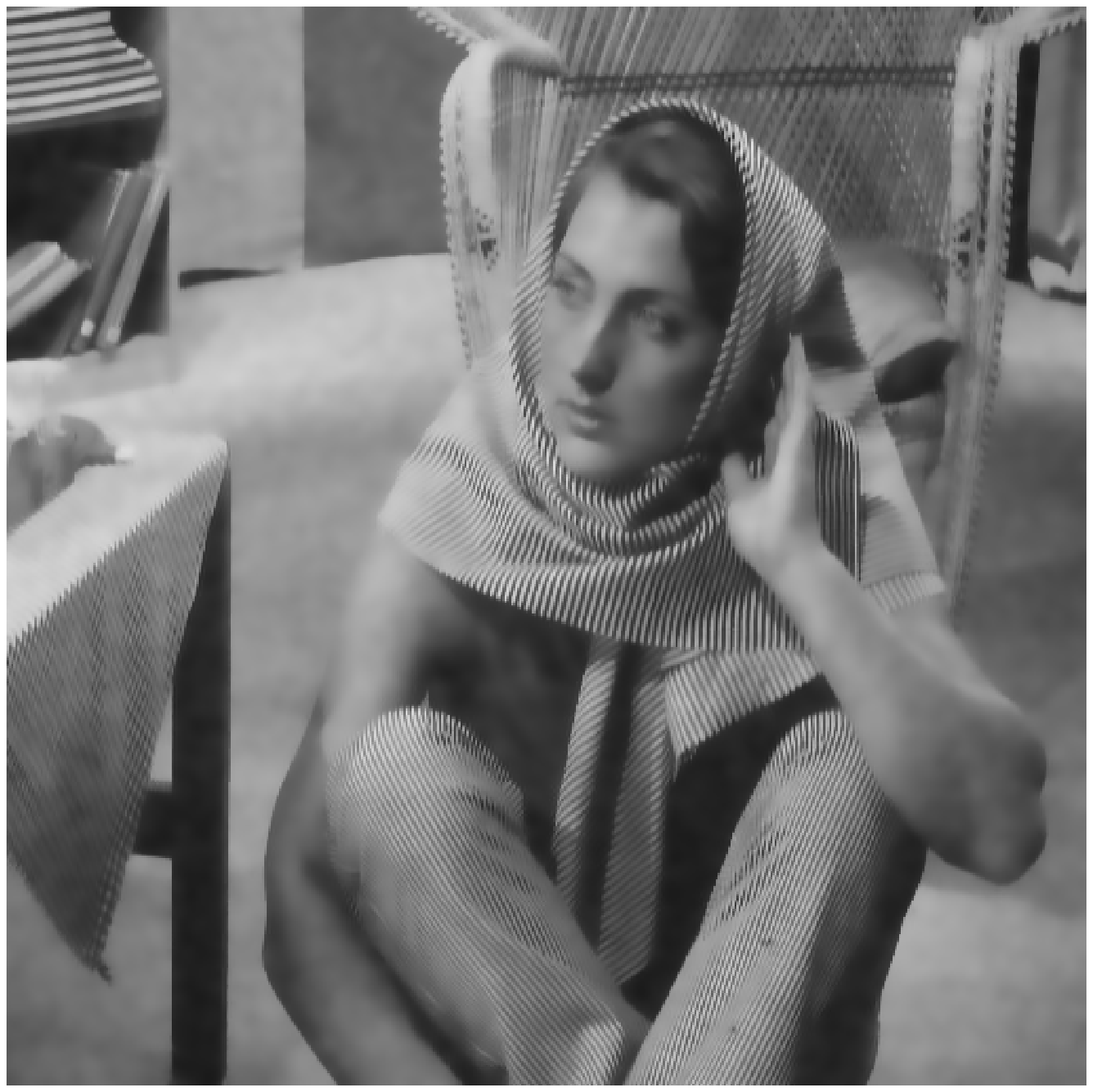}} 
\caption{Standard versus oracle bilateral filter. The standard and the oracle filters respectively use the corrupted and the clean image to compute the range filter. Notice that the PSNR improves by more than $3$ dB, and the image from the oracle looks visibly much better than that obtained using the standard filter. The result in (d) is the ``best'' one can hope to achieve if, instead of the clean image, some proxy is used  to compute the range filter.} 
\label{OracleBF}
\end{figure*}

The bilateral filter was proposed by Tomasi and Maduchi \cite{Tomasi1998} as a simple, non-iterative alternative to the
Perona-Malik diffusion. The origins of the filter can be traced back to the work of Lee \cite{Lee1983} and Yaroslavsky \cite{Yaroslavsky1985}. The SUSAN framework of Smith and Brady \cite{Smith1997} is also based 
on a similar idea. 

For a discrete image $( f(\i) )_{\i \in \Z^2}$, the  bilateral filter is given by 
\begin{equation}
\label{BF}
 f_{\text{BF}}(\i)=  \frac{\sum_{\j \in \Omega} g_{\sigma_s}(\j) \  g_{\sigma_r}(f(\i-\j)-f(\i)) \ f(\i-\j)}{\sum_{\j \in \Omega} g_{\sigma_s}(\j)  \  g_{\sigma_r}(f(\i-\j)-f(\i)) }.
\end{equation}
In this formula, $g_{\sigma_s}(\i)$ is the \textit{spatial} filter defined on some neighbourhood $\Omega$ and $g_{\sigma_r}(t)$ is the \textit{range} filter. Typically, $\Omega$ is a square neighbourhood, $\Omega=[-W,W] \times [-W,W]$, and  both the spatial and range filters are Gaussian:
\begin{equation*}
g_{\sigma_s}(\i) = \exp\left(- \frac{\lVert \i \rVert^2}{2\sigma_s^2}\right) \quad \text{and} \quad g_{\sigma_r}(t) = \exp\left(- \frac{t^2}{2\sigma_r^2}\right).
\end{equation*}
The support $W$ of the spatial filter is usually set to be $3\sigma_s$.

The spatial filter puts larger weights on pixels that are close to the pixel of interest compared to distant pixels.
On the other hand, the range filter operates on the intensity differences between the pixel of interest  and its neighbors (which makes the overall filter non-linear).  
The role of the range filter is to restrict the averaging to neighbouring pixels whose intensities are similar to that of the pixel of interest.
In particular, $f(\i-\j)-f(\j)$ in \eqref{BF} is close to zero when both $\i$ and its neighbors $\{\i-\j : \j \in \Omega\}$ belong to a homogenous region. In this case, $g_{\sigma_r}(f(\i-\j)-f(\i)) \approx 1$, and \eqref{BF} effectively acts as a standard Gaussian filter. On the other hand, consider the situation in which the pixel of interest $\i$ is in the vicinity of an edge. If $\i - \j$ and $\i$ are on the opposite sides of the edge, then $g_{\sigma_r}(f(\i-\j)-f(\i))$ is relatively small compared to what it is  when $\i - \j$ and $\i$  are on the same  side of the edge. 
This effectively prohibits the mixing of pixels from different sides of the edge during the averaging, and hence avoids the blurring that is otherwise induced by linear filters.

\subsection{Present Contribution}

The bilateral filter has found widespread applications in  image processing, computer vision, and  computational photography. We refer the interested reader to \cite{Book2009} and the references therein for an exhaustive account of various applications. 

Our present interest is in the image denoising applications of the filter \cite{Elad2002,Aleksic2006,Liu2006}. 
The bilateral filter has received renewed attention in the image processing community in the context of image denoising \cite{Knaus2014,Morel2014}. 
It is well-known that, while the filter is quite effective in removing modest amounts of additive noise, its denoising performance is severely impaired at large noise levels \cite{Book2009,Buades2005}. To overcome this drawback, different iterative forms of the filter were proposed in \cite{Elad2002}, for example. In a different direction, it was shown by Buades et al. \cite{Buades2005} that a patch-based extension of the filter can be used to bring the denoising performance of the filter at par with state-of-the-art methods. However, this and other advanced patch-based methods \cite{Kervrann2006,KSVD,BM3D} are much more computation-intensive than the bilateral filter.

In this paper, we demonstrate how the denoising performance of the bilateral filter can be improved at almost no additional cost by incorporating a simple pre-processing step into the framework. 
To the best of our knowledge, this improvement has not been reported in the literature on bilateral filtering-based denoising. 
Although the present improvement is not of the order of the improvement provided by K-SVD and BM3D, we will demonstrate that the improved bilateral filter is often competitive with the Non-Local Means (NLM) filter \cite{Buades2005}, while being significantly cheaper. Moreover, we also describe a fast algorithm for the modified filter which should be of interest in real-time denoising. 

\subsection{Organization}

The rest of the paper is organized as follows. In Section \ref{sec:IBF}, we briefly describe the denoising problem and the standard metrics that are used to quantify the denoising performance. We then introduce the proposed improvement of the bilateral filter in the context of denoising. In this section, we also report a fast algorithm for the proposed filtering. Experimental results on synthetic and natural images are provided in Section \ref{sec:experiments}, and we conclude the paper in Section \ref{sec:conclusion}.

\section{Improved Bilateral Filter}
\label{sec:IBF}

We consider the problem of denoising grayscale images that are corrupted with additive white Gaussian noise. 
In this setup, we are given the \textit{corrupted}  (or \textit{noisy}) image
\begin{equation}
\label{noise}
f(\i) = f_0(\i) + \sigma \cdot n_{\i} \qquad (\i \in I),
\end{equation}
where 
\begin{itemize}
\item $ I$ is some finite rectangular domain of $\Z^2$, 
\item $(f_0(\i))_{\i \in I}$ is the unknown \textit{clean} image, and
\item $(n_{\i})_{\i \in I}$ are independent and distributed as $\mathcal{N}(0,1)$.
\end{itemize}
The goal is find a \textit{denoised} estimate $\hat{f}(\i)$ of the clean image from the corrupted samples. The denoised image should visually resemble the clean image. To quantify the resemblance, two standard metrics are widely used in the image processing literature, namely the peak signal-to-noise ratio (PSNR) and the structural similarity index (SSIM) \cite{SSIM2004}. The PSNR is defined to be  $10 \log_{10}(255^2/\text{MSE})$, where $$\text{MSE}= \frac{1}{|I|} \sum_{\i \in I} (\hat{f}(\i) - f_0(\i))^2.$$

\subsection{Proposed Improvement}

In linear diffusion, the clean image is estimated by linearly averaging the noisy samples. 
The averaging process successfully brings down the noise floor in homogenous regions by a factor of $\sqrt{L}$, where $L$ is the length of the filter \cite{Buades2005}. 
However, the filter also implicitly acts on  the underlying clean image in the process. As a result, it introduces blurring in the image features besides reducing the noise floor.
This can precisely be overcome by applying the bilateral filter on the corrupted image \cite{Elad2002,Aleksic2006,Liu2006}. In this regard, note that the range filter operates on the noisy samples. In other words, the corrupted image is used not just for the averaging but also to control the blurring via the range filter. What if the range filter could directly operate on the clean image? That is, instead of \eqref{BF}, suppose we consider the formula
\begin{equation}
\label{OBF}
 f_{\text{OBF}}(\i)=  \frac{\sum_{\j \in \Omega} g_{\sigma_s}(\j) \  g_{\sigma_r}(f_0(\i-\j)-f_0(\i)) \ f(\i-\j)}{\sum_{\j \in \Omega} g_{\sigma_s}(\j)  \  g_{\sigma_r}(f_0(\i-\j)-f_0(\i)) }.
\end{equation}
The denoising result obtained using this ``oracle'' filter is compared with that obtained using \eqref{BF} in Figure \ref{OracleBF}. It is not surprising that the result obtained using the oracle filter is visibly better and has higher PSNR.

Of course, the problem is that we do not have access to the clean image in practice, and thus the oracle bilateral filter cannot be realized.
Nevertheless, one could consider some form of proxy for the clean image. For example, one could use an ``iterated'' bilateral filter \cite{Elad2002} where the output of the bilateral filter is used as a proxy. However, this requires us to compute \eqref{BF} twice, which doubles the run time of the filter. Our present proposal is simply to use the box-filtered version of the noisy image as a proxy. In other words, the proposed improved bilateral filter (in short, IBF) is given by
\begin{equation}
\label{IBF}
 f_{\text{IBF}}(\i)=  \frac{\sum_{\j \in \Omega} g_{\sigma_s}(\j) \  g_{\sigma_r}(\bar{f}(\i-\j)- \bar{f}(\i)) \ f(\i-\j)}{\sum_{\j \in \Omega} g_{\sigma_s}(\j)  \  g_{\sigma_r}(\bar{f}(\i-\j)-\bar{f}(\i)) }.
\end{equation}
where
\begin{equation}
\label{box}
 \bar{f}(\i)= \frac{1}{(2L+1)^2} \sum_{\j \in [-L,L]^2} \!\!\! f(\i-\j).
\end{equation}
Clearly, the amount of smoothing induced by the box-filter is controlled by $L$. When $L$ is very small, $ \bar{f}(\i) \approx f(i)$, and \eqref{IBF} behaves as \eqref{BF}. At the other other end, the image structures are over-smoothed when $L$ is large and this makes $\bar{f}(\i)$ a bad proxy for the original image. 
Thus, $L$ should not be too small and neither too large. We will report the appropriate choice of $L$ in the sequel.
\begin{figure*}[!htp]
\centering
\includegraphics[width=1.0\linewidth]{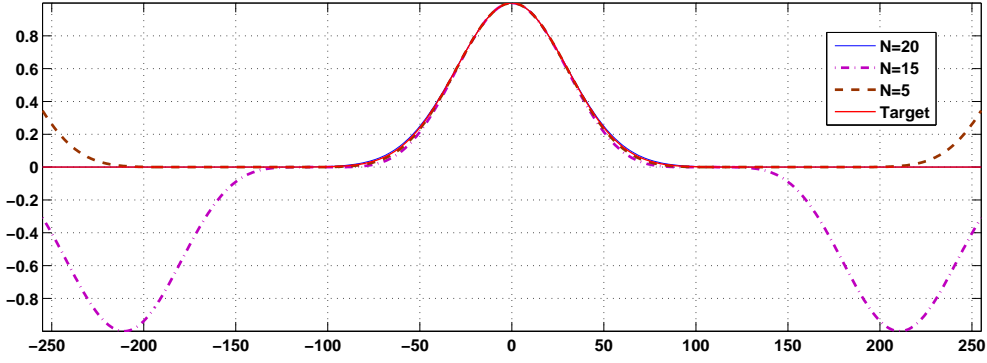}
\caption{Approximation of the range kernel $g_{\sigma_r}(t)$ on the interval $[-255,255]$ using raised-cosines of different orders. In this case, $\sigma_r=30$ .
Note that the approximation around the origin is already good when $N=5$, but the approximation at the boundaries is far from satisfactory (due to the periodic nature of the raised-cosine). 
Higher-order raised-cosines are required to suppress the oscillation at the boundaries and to force non-negativity and monotonicity.} 
\label{approximation}
\end{figure*}

\subsection{Fast Implementation}
\label{sec:fast}

The cost of computing \eqref{IBF} is almost identical to that of computing \eqref{BF}, since the additional cost of computing \eqref{box} is negligible in comparison. More precisely, the cost of computing \eqref{IBF} is $O(|\sigma_s^2|)$ per pixel, since the support $\Omega$ of the spatial  filter is proportional to $\sigma_s^2$. On the other hand, it is well-known that the box-filter in \eqref{box} can be computed using $O(1)$ operations per pixel  \cite{Deriche1993}. We now explain how we can implement \eqref{IBF} using $O(1)$ operations (with respect to $\sigma_s$) as a straightforward extension of the algorithm proposed in \cite{Chaudhury2011,Chaudhury2013}. 

For completeness, we present the main ideas behind the fast algorithm in \cite{Chaudhury2011}. Note that the effective domain of the range filter in \eqref{IBF} is the interval $[-T,T]$, where  $T$ is the maximum value of  $\bar{f}(\i-\j) - \bar{f}(\i)$ over all $\i \in I$ and $\j \in [-W,W]^2$. In other words, $T$ is the maximum ``local'' dynamic range of $\bar{f}(\i)$ over square boxes of length $2W$. 
Note that the complexity of computing $T$ is comparable to that computing the filter, namely $O(W^2)$ operations per pixel.
A fast algorithm for computing $T$ was however later proposed in \cite{Chaudhury2013}, which we will use in this paper.

It was observed in \cite{Chaudhury2011} that
\begin{equation}
\label{approx}
g_{\sigma_r}(t) = \lim_{N \rightarrow \infty} \ \left[ \cos\left( \frac{t}{\sigma_r \sqrt{N}} \right) \right]^N.
\end{equation}
The functions on the right are called \textit{raised cosines}, and we refer to $N$ as its order. Note that while \eqref{approx} holds for every $t$, we only require a good approximation for $t \in [-T,T]$. Moreover, the raised cosine should ideally be positive and monotonic on this interval. In particular, one can verify that if $N$ is larger than  $N_0 = 0. 405 (T/\sigma_r)^2$, then the raised-cosine
is positive and monotonic on $[-T,T]$.  In other words, any raised-cosine of order $N \geq N_0$ is an acceptable approximation of the Gaussian range filter. 
The approximation process in demonstrated in Figure \ref{approximation}.

Now, using the identity $2\cos \theta = e^{\imath \theta} + e^{-\imath \theta}$ (where $\imath$ denotes the imaginary unit $\sqrt{-1}$) and the binomial expansion, we can write 
\begin{eqnarray}
\label{shiftability}
\begin{aligned}
\left[ \cos\left( \frac{t}{\sigma_r \sqrt{N}} \right) \right]^N= \sum_{n=0}^N c_n  \exp \left(\imath \omega_n t \right),
\end{aligned}
\end{eqnarray}
where 
\begin{equation}
\label{coeff}
c_n = \frac{1}{2^N}\binom{N}{n} \quad \text{and} \quad \omega_n=\frac{(2n-N)}{\sigma_r \sqrt{N}}.
\end{equation}
%
%
%
Plugging \eqref{shiftability} into \eqref{IBF}, using the multiplication-addition property $\exp(u + v)=\exp(u)\exp(v)$, and exchanging the summations, we can express the numerator of \eqref{IBF} as
\begin{eqnarray}
\label{num}
\begin{aligned}
\sum_{n=0}^N  c_n  \exp \left(-\imath \omega_n \bar{f}(\i)\right)  (F_n \ast g_{\sigma_s})(\i),
\end{aligned}
\end{eqnarray}
where 
$$F_n(\i) = \exp(\imath \omega_n \bar{f}(\i)) f(\i),$$ 
and  
\begin{equation*}
(F_n \ast g_{\sigma_s})(\i) = \sum_{\j \in \Omega} g_{\sigma_s}(\j) F_n(\i-\j)
\end{equation*}
denotes the Gaussian filtering of $F_n(\i)$. Similarly, we have the following approximation for the denominator of \eqref{IBF}: 
\begin{equation}
\label{denom}
\sum_{n=0}^N  c_n  \exp \left(-\imath \omega_n \bar{f}(\i)\right)  (G_n \ast g_{\sigma_s})(\i),
\end{equation}
where $$G_n(\i) = \exp(\imath \omega_n \bar{f}(\i)).$$

To summarize then, by using the raised-cosine approximation of the Gaussian range filter, we can express the numerator and denominator of \eqref{IBF} as a linear combinations of Gaussian filters applied on the images $F_n(\i)$ and $G_n(\i)$.  It is well-known that Gaussian filtering can be computed using $O(1)$ operations per pixel (i.e., independent of $\sigma_s$) using recursions \cite{Deriche1993}. 
As a result, we can compute \eqref{num} and \eqref{denom}, and hence the overall filter \eqref{IBF}, using $O(1)$ operations per pixel.

In fact, it is further possible to cut down the run time without appreciably sacrificing the approximation using truncation \cite{Chaudhury2013}. In particular, it can be verified that the contribution of the central  terms in \eqref{shiftability} to the overall approximation is less compared to the other terms. Indeed, the distribution of the normalized binomial coefficients $(c_n)$ is bell-shaped with a peak around $N/2$.  As a result, one can truncate the sum away from the central peak and tradeoff speed versus accuracy. In particular, given some tolerance $\varepsilon >0$, we incrementally  find the largest $M$ such that 
\begin{equation}
\label{trunc}
c_M + \ldots + c_{N - M} > 1- \varepsilon/2.
\end{equation}
We can then further approximate \eqref{shiftability} using the truncated sum 
\begin{equation}
\label{approx1}
\sum_{n=M}^{N-M} c_n  \exp \left(\imath \omega_n t \right).
\end{equation}
Note that the error between \eqref{approx1}  and \eqref{shiftability} is
\begin{equation*}
\sum_{n=0}^{N} c_n  \exp \left(\imath \omega_n t \right) - \sum_{n=M}^{N-M} c_n  \exp \left(\imath \omega_n t \right),
\end{equation*}
whose magnitude is, by construction, at most as large as $2(c_0+\ldots+c_M) \leq \varepsilon$. Using \eqref{approx1}, we can further cut down the run time of \eqref{IBF} by a factor of about $2M/N$, without any appreciable change in denoising performance. For large $N$, one can test for the condition $c_M + \ldots + c_{N - M} > 1- \varepsilon/2$ without having to compute all the $c_n$. In particular, the Chernoff-inequality for the binomial distribution gives us the estimate 
\begin{equation}
\label{Chernoff}
 M = \frac{1}{2}\left(N - \sqrt{4N\log(2/\varepsilon)}\right),
\end{equation}
which is quite accurate when $N>100$  \cite{Chaudhury2013}. 

The complete algorithm is summarized in Algorithm \ref{algoIBF}. Here, we have used $[N]$ to denote the set of numbers $0,1,\ldots,N$, and $[N/M]$ to denote the set of numbers  $0,\ldots,M$ and $N-M,\ldots,N$.
In  step $6$ (c), we have used $G^{\ast}_n(\i)$ to denote the complex conjugate of $G_n(\i)$.

\begin{algorithm*}
\KwData{Corrupted image $f(\i)$, and filter parameters $\sigma_s,\sigma_r,\varepsilon$.}
\KwResult{Filtered image $f_{\text{IBF}}(\i)$.}
\textbf{Initialize}: Images $P(\i)=0$ and $Q(\i)=0$ for all $\i$\;
Compute box-filtered image $\bar{f}(\i)$ using \eqref{box}\;
Compute local dynamic range $T$ of $\bar{f}(\i)$\;
$N = 0. 405 (T/\sigma_r)^2$\;
\Switch{}{
\Case{$N < 40$}{
$M=0$\;
 Set $c_n,\omega_n$ for $n \in [N]$ using \eqref{coeff}\;
}
\Case{$40 \leq N < 100$}{
 Set $c_n,\omega_n$ for $n \in [N]$ using \eqref{coeff}\;
 Fix $\varepsilon=0.01$ and find largest $M$ such that \eqref{trunc} holds\;
}
\Case{$N \geq 100$}{
Set $M$ using \eqref{Chernoff} with $\varepsilon=0.1$\;
 Set $c_n,\omega_n$ for $n \in [N/M]$ using \eqref{coeff}\;
}
}
 \For{$n=M,\ldots,N-M$}{
$G(\i) =\exp(\imath \omega_n \bar{f}(\i))$\;
$F(\i) =G(\i) f(\i)$\;
$H(\i)=c_n G^{\ast}(\i)$\;
$P(\i)=P(\i) + H(\i) \cdot (F \ast g_{\sigma_s} )(\i)$\;
$Q(\i)=Q(\i)+ H(\i) \cdot  (G \ast g_{\sigma_s})(\i)$\;
}
$f_{\text{IBF}}(\i)=P(\i)/Q(\i)$ for all $\i$\;
\caption{Fast Improved Bilateral Filtering}
\label{algoIBF}
\end{algorithm*}

\section{Experiments}
\label{sec:experiments}

\subsection{Complexity and Run Time}

The complexity of the direct implementation of the proposed filter is identical to that of the standard bilateral filter, namely $O(\sigma_s^2)$.
On the other hand,  for an image with maximum local dynamic range $T$, the complexity of the fast implementation proposed in Section \ref{sec:fast} is $O(T^2/\sigma^2_r)$. The final run time is however determined by the constants that are implicit in the above complexity estimates. 
In practice, the main speed-up is due to the fact that Gaussian filtering can be implemented very efficiently using standard packages. For example, the image filtering in step $6$ (d) and (e) can be done using the optimized ``imfilter'' routine in Matlab.
In Table \ref{table1}, we compare the run times of the fast implementation and the direct implementation (for typical filter parameters) on the \textit{Barbara} image. 

\begin{table}[!htb]
\caption{Comparison of the average run times of the direct and the fast implementation of the proposed filter  for different $(\sigma_s,\sigma_r)$. We used  \textit{Barbara} for the comparison and the noise level was set at $\sigma=20$. 
For both implementations, we took the support of the spatial Gaussian to be $W=3\sigma_s$.}  
\vspace{2mm}
\centering 
\begin{tabular}{l*{6}{c}r}
\hline
     &       $(2,15)$  & (4,20) & $(3,25)$  & $(5,30)$ & $(3,35)$  & $(4,40)$ \\
\hline
Direct          & 16.5s     & 60.5s   & 35.3s     &93.8s    &35.5s   & 60.5s   \\
Fast            & 0.52s         &0.61s    & 0.52s  &0.47s      &0.43s       &0.47s  \\
\hline
\end{tabular}
\label{table1}
\end{table}

All computations were performed using  Matlab on a quad-core 2.70 GHz Intel machine with 16 GB memory.
It is clear from the table that a significant acceleration is achieved using the fast algorithm. 
In particular, notice that the fast implementation takes about $0.5$ seconds for different parameter settings. 
On the other hand, notice the run time of the direct implementation scales up quickly with the increase in the width of the spatial filter.
We note that the run time of the fast  implementation  can be further cut down using a parallel (multithreaded) implementation of  step $6$ in Algorithm \ref{algoIBF}.

\subsection{Optimal Choice of $L$}

\begin{figure*}
\centering
\subfloat[\textit{Barbara}.]{\includegraphics[width=0.33\linewidth]{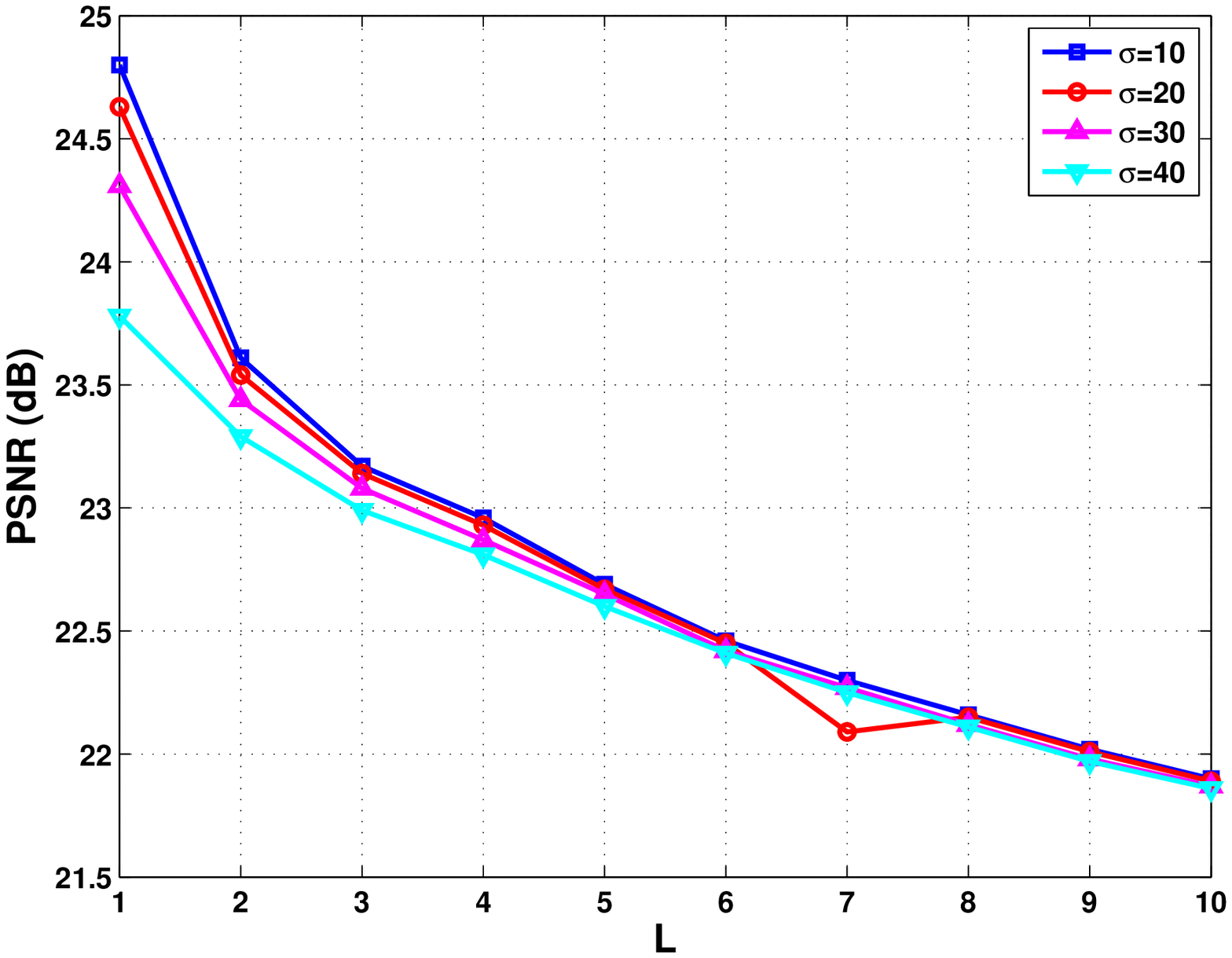}}  
\subfloat[\textit{Lena}.]{\includegraphics[width=0.33\linewidth]{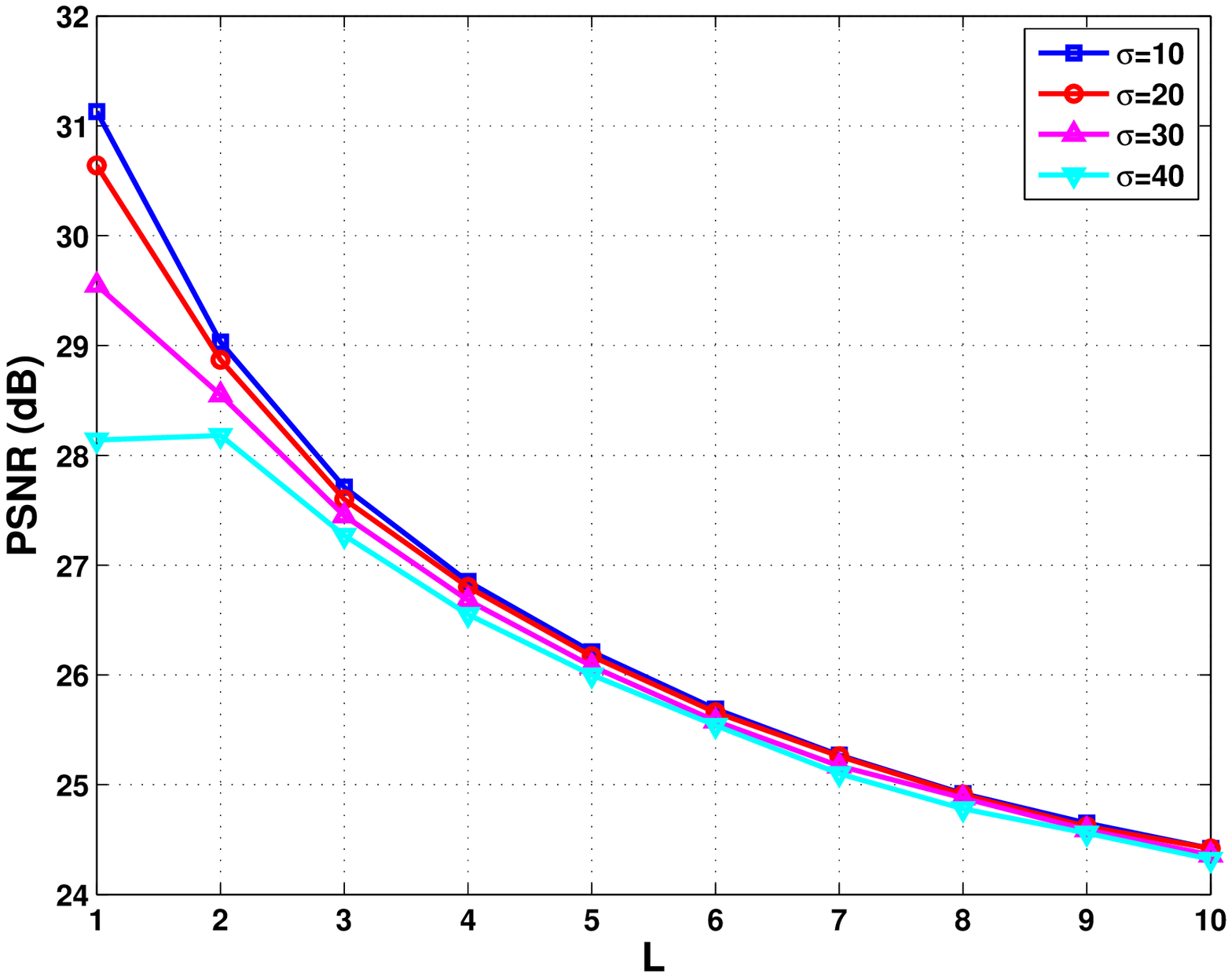}}   
\subfloat[\textit{House}.]{\includegraphics[width=0.33\linewidth]{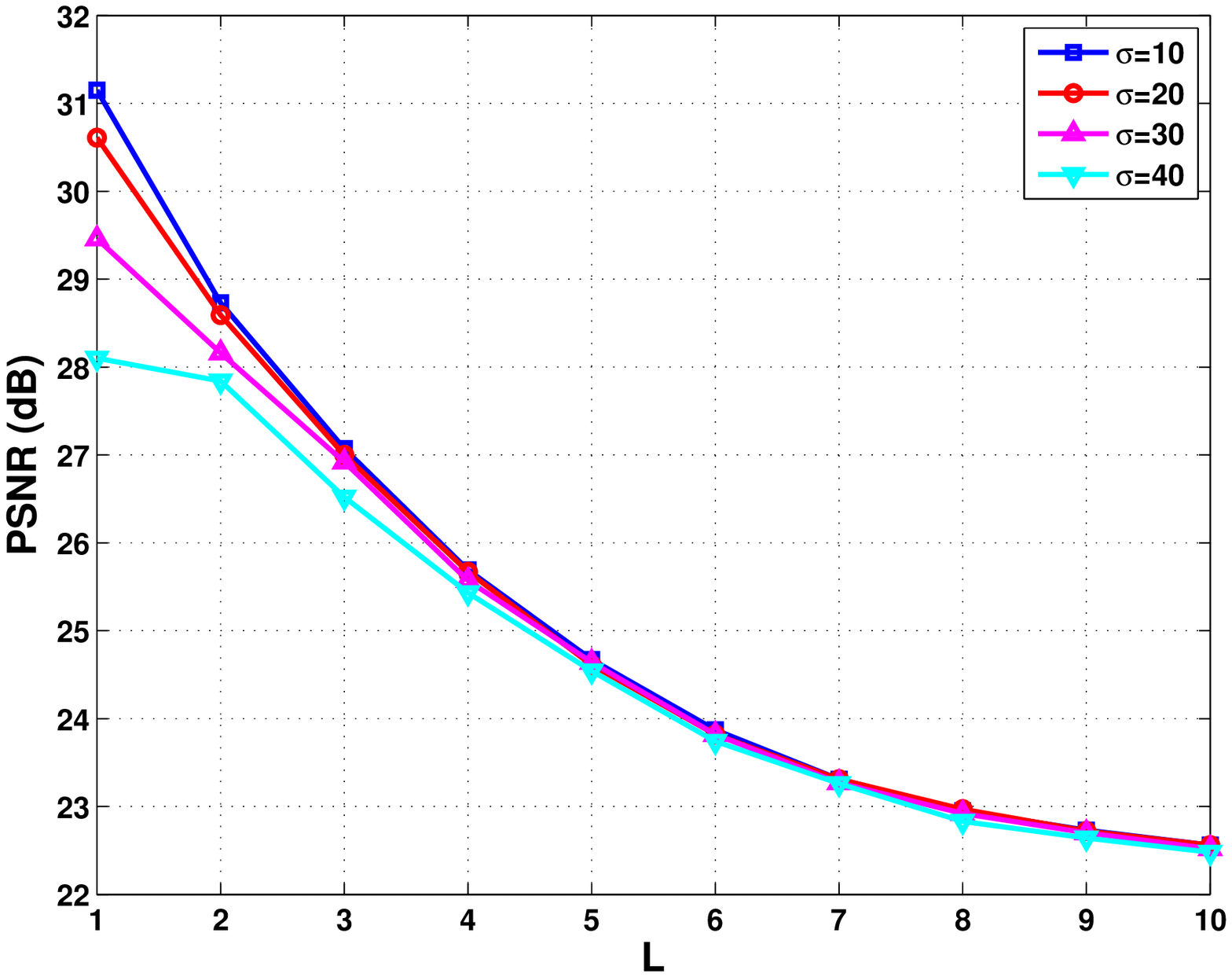}} 
\caption{The plots show how the PSNR varies with $L$ at various noise levels. For all the experiments, we used $\sigma_s=6$ and $\sigma_r=20$. Notice that the PSNR is consistently highest when $L=1$. } 
\label{PSNRvsL}
\end{figure*}

We now come to question about the choice of the optimal length $L$ for the box-filter in \eqref{box}. We performed exhaustive some simulations in this direction, the results of some of which are reported in Figure \ref{PSNRvsL}. For these simulations, we conclude that a box-filter with $L=1$ ($3 \times 3$ blur) is optimal for most settings. A possible way to explain this is that this small filter is able to suppress the noise without excessively blurring the image features.

\subsection{Denoising Results}

\begin{figure*}
\centering
\subfloat[Corrupted ($\sigma=40, 16.12$ dB).]{\includegraphics[width=0.25\linewidth]{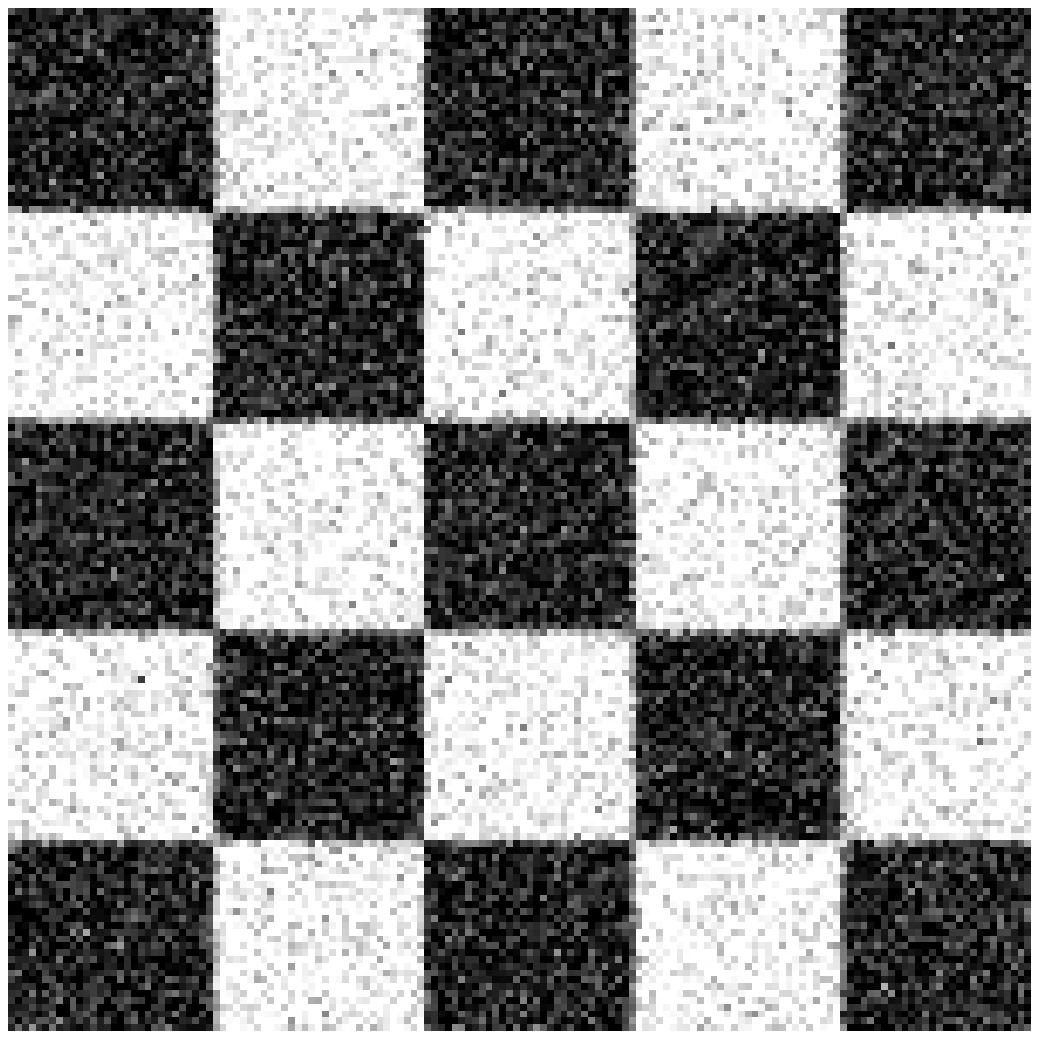}}  
\subfloat[Standard Bilateral ($18.92$ dB).]{\includegraphics[width=0.25\linewidth]{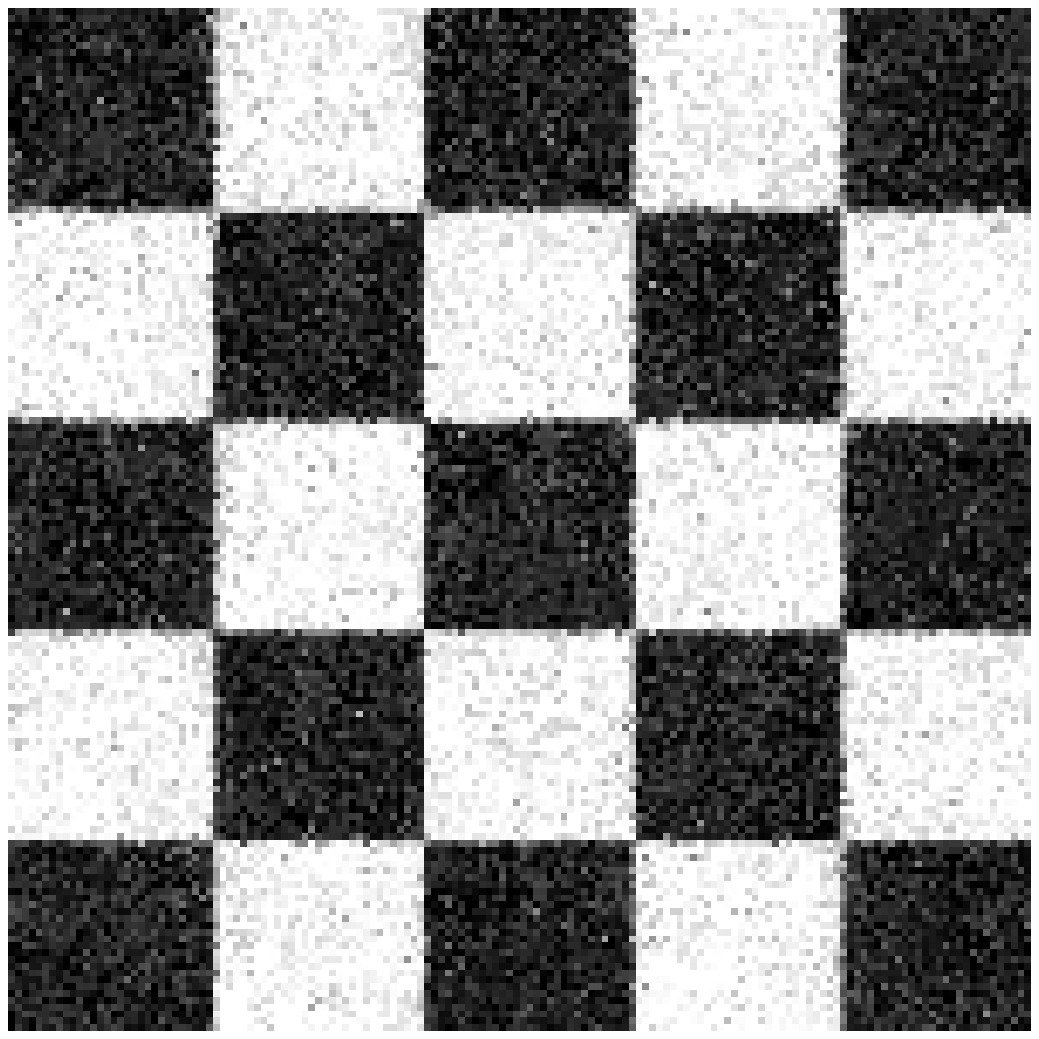}}   
\subfloat[Proposed ($\bf{29.23}$ dB).]{\includegraphics[width=0.25\linewidth]{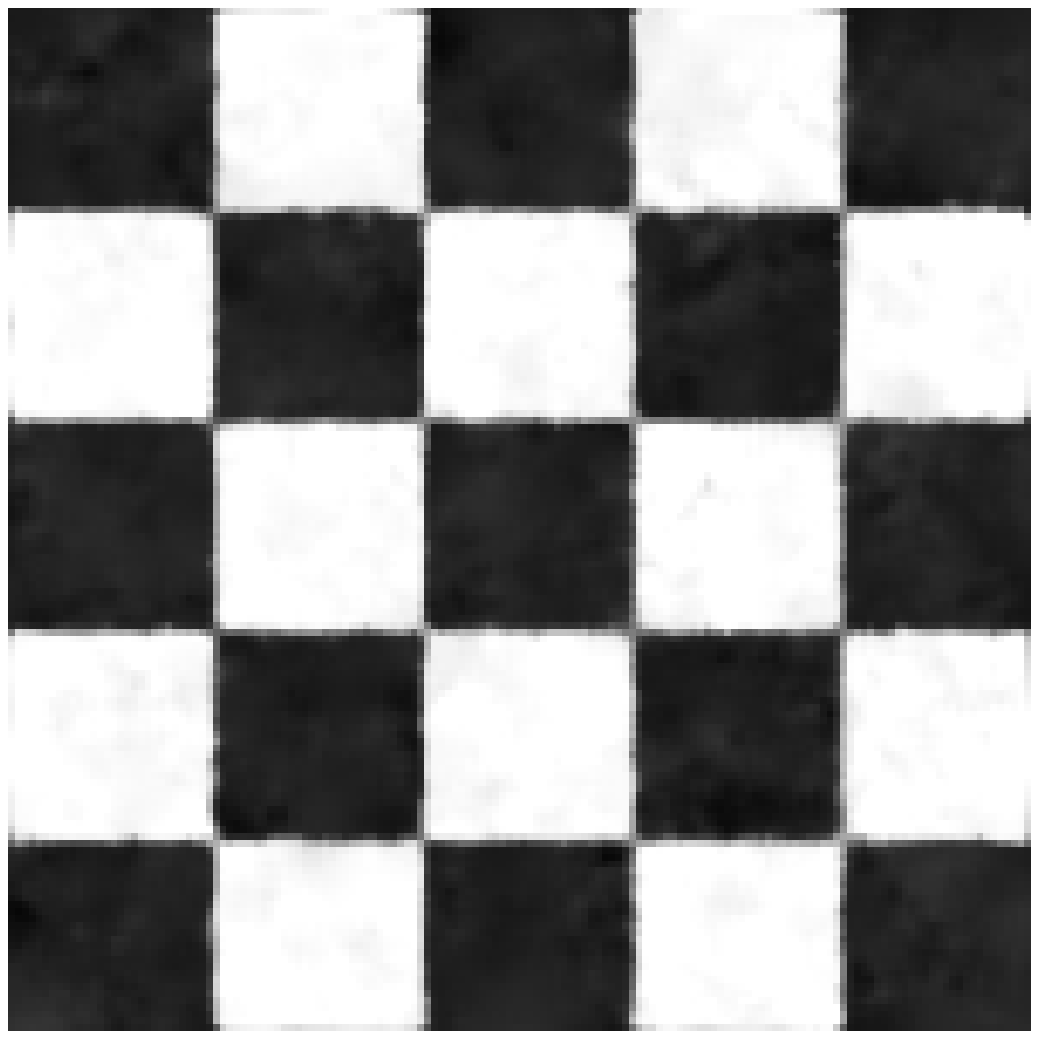}} 
\subfloat[Oracle Bilateral ($32.16$ dB).]{\includegraphics[width=0.25\linewidth]{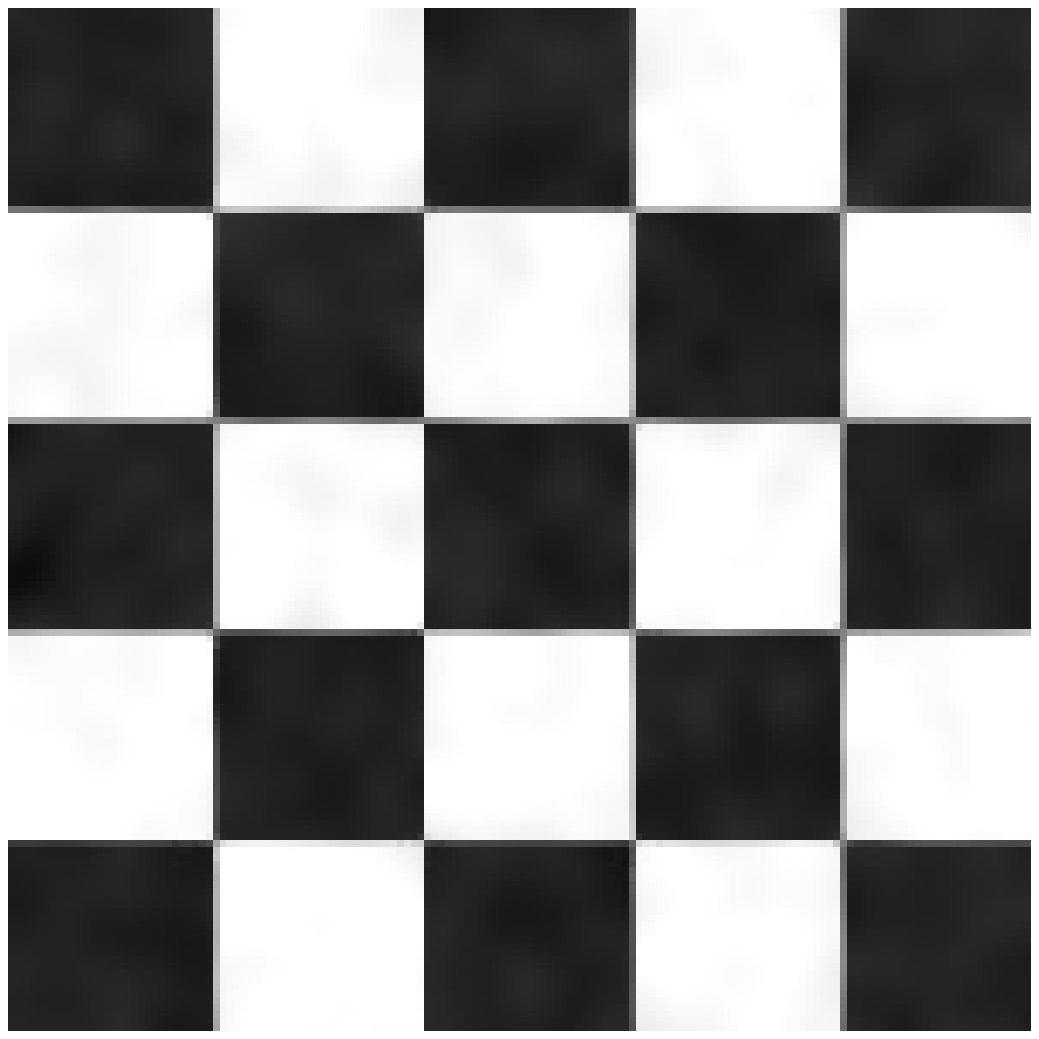}} 
\caption{Denoising results for \textit{Checker} (original size $150 \times 150$). Parameters used: $\sigma_s=3,\sigma_r=30, \varepsilon=0.01$, and $W=3\sigma_s$. Notice that the image in (c) looks significantly better than that in (b) and is similar to the oracle in (d). } 
\label{ckb}
\end{figure*}

\begin{figure*}
\centering
\subfloat[Corrupted ($\sigma=25, 20.13$ dB).]{\includegraphics[width=0.25\linewidth]{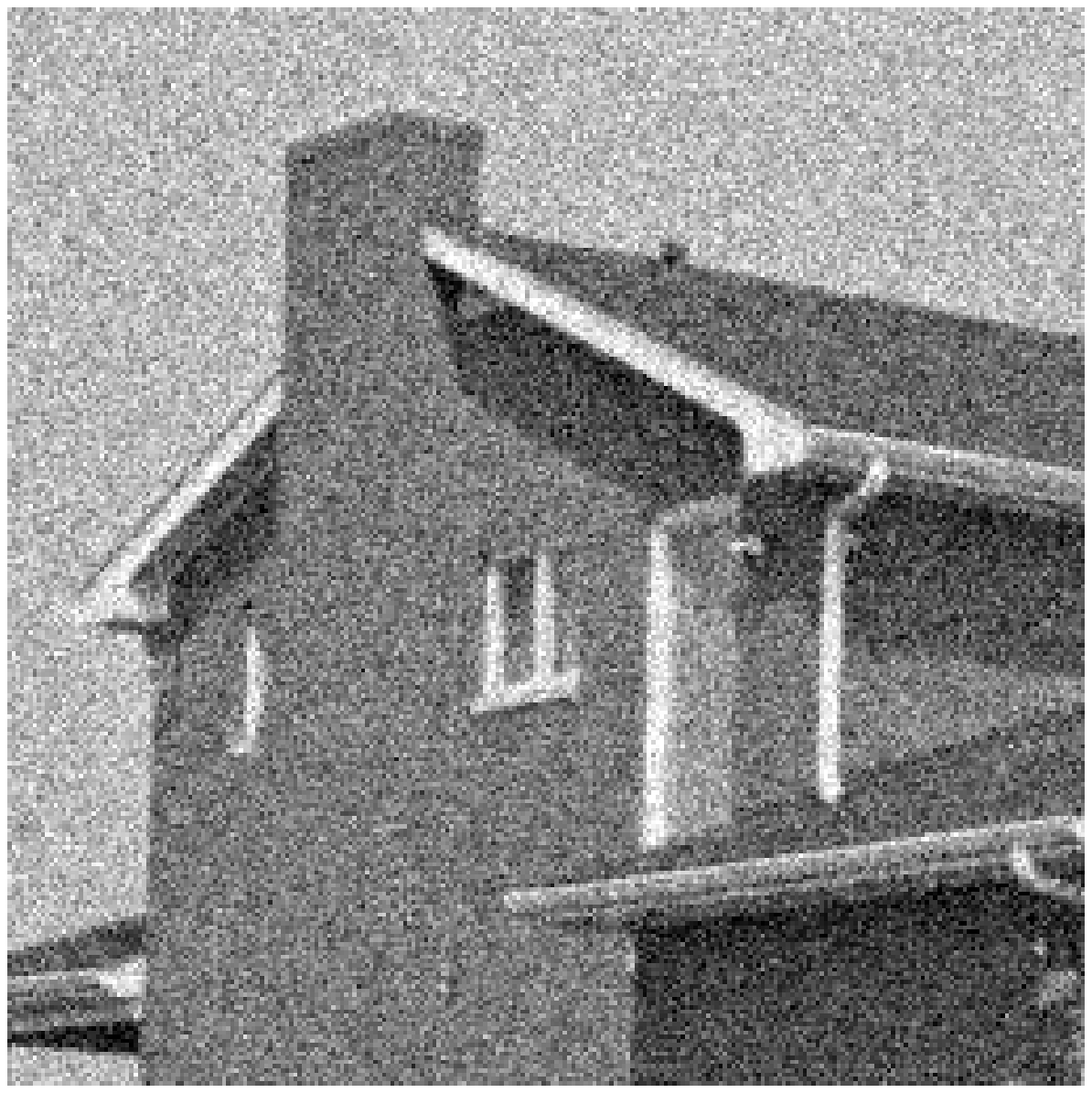}}  
\subfloat[Standard Bilateral ($25.52$ dB).]{\includegraphics[width=0.25\linewidth]{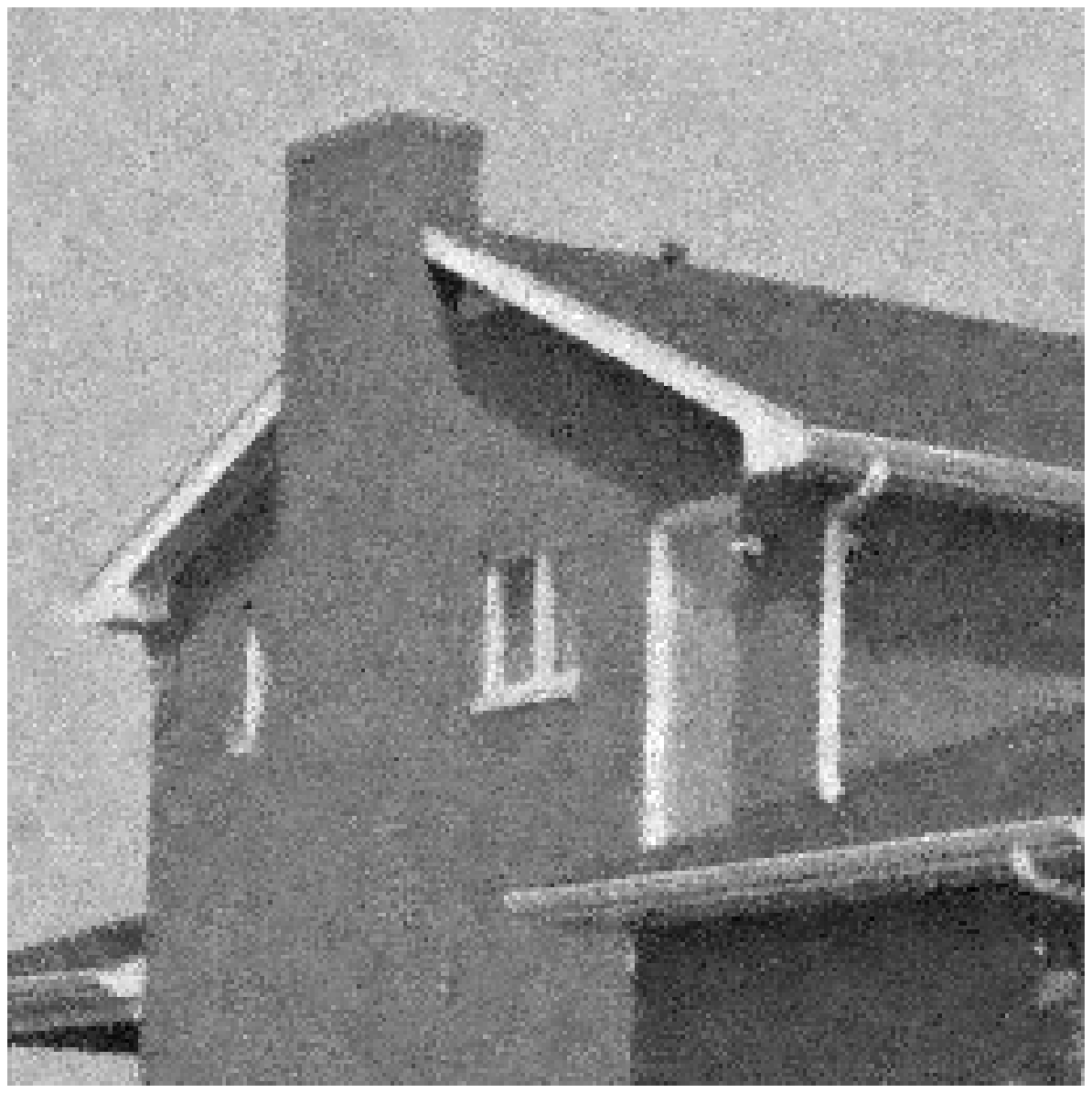}}   
\subfloat[Proposed ($\bf{29.92}$ dB).]{\includegraphics[width=0.25\linewidth]{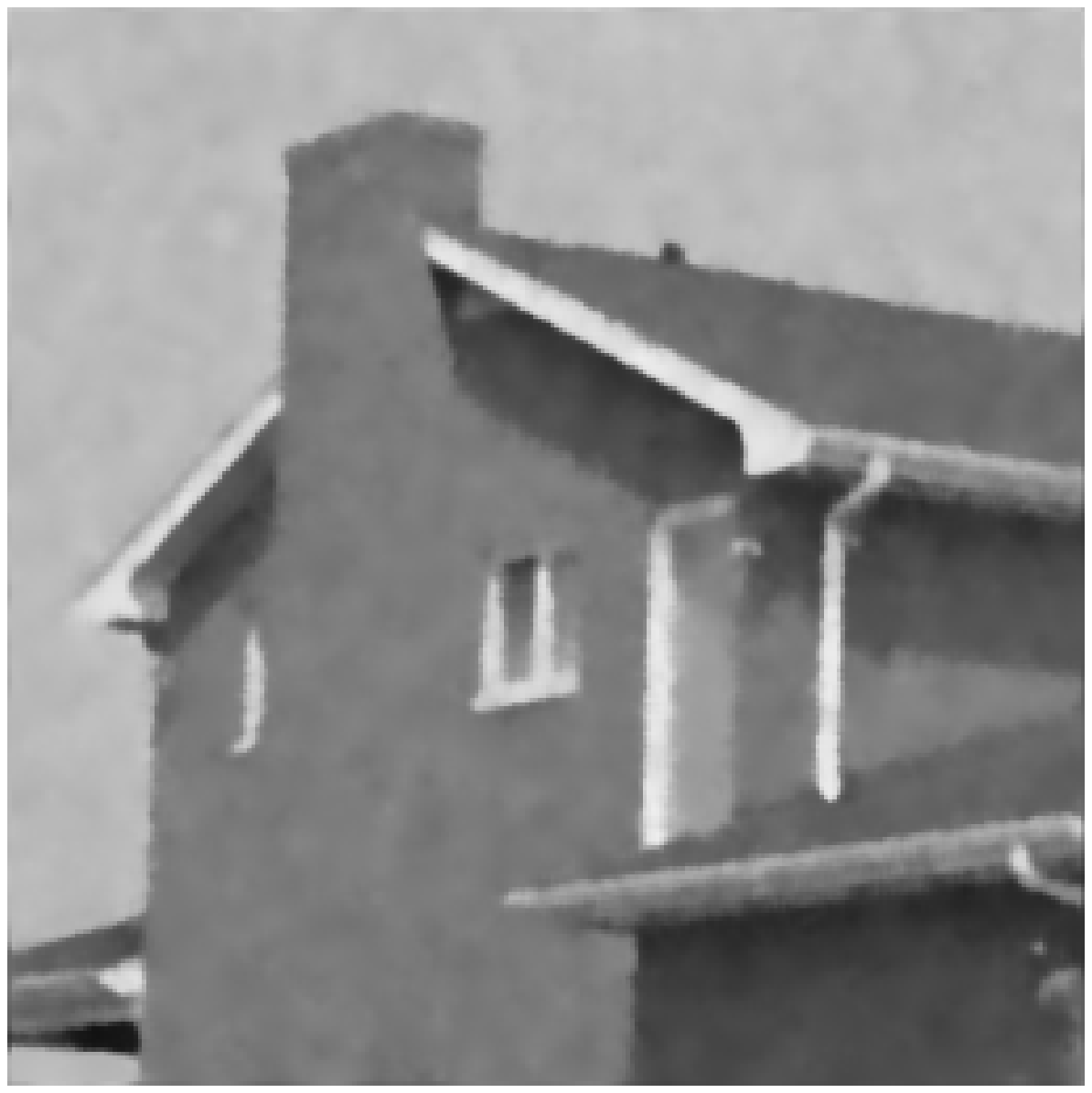}} 
\subfloat[Iterated Bilateral ($28.71$ dB).]{\includegraphics[width=0.25\linewidth]{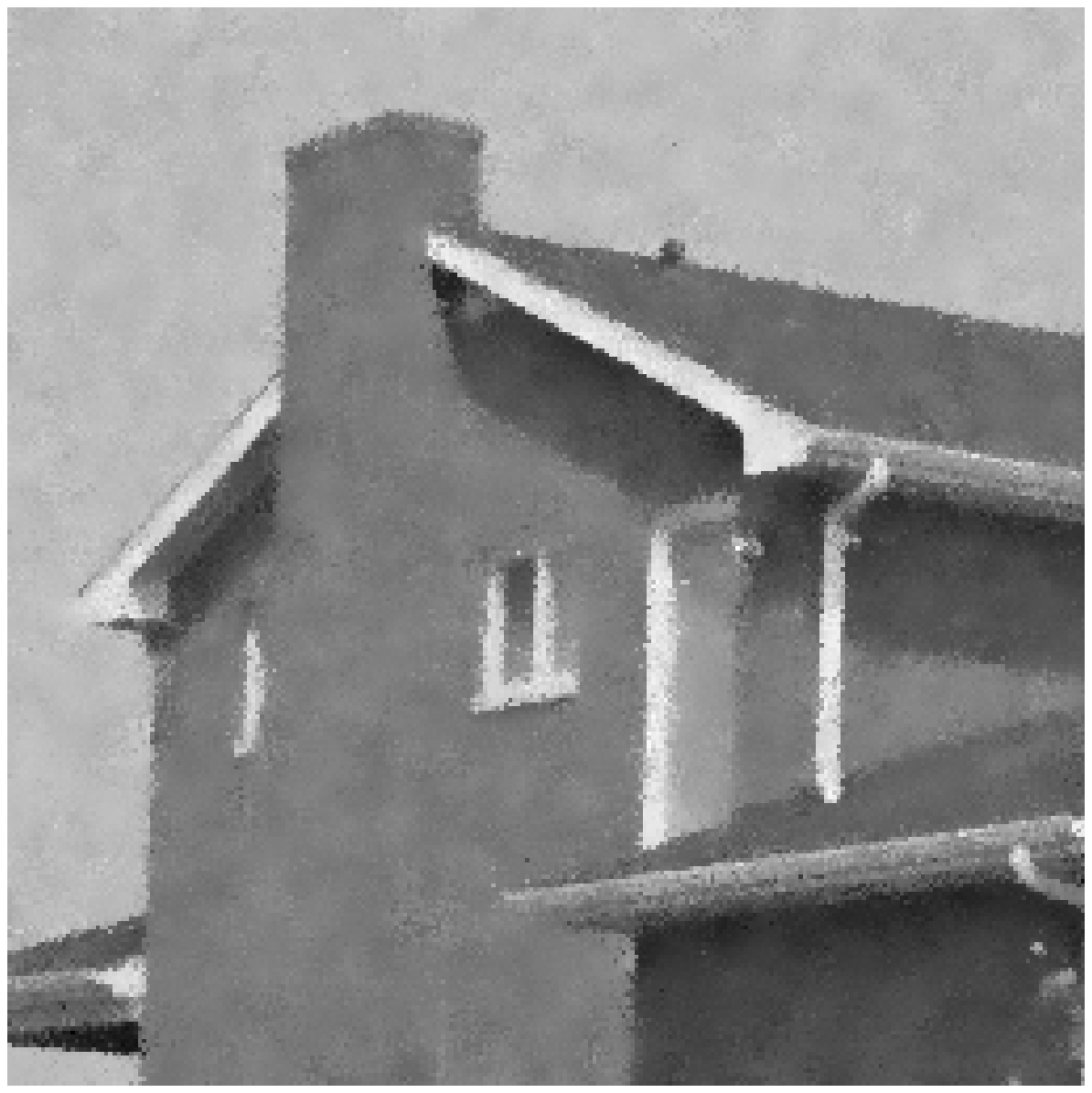}} 
\caption{Denoising results for \textit{House}. The filter parameters in this experiment are identical to the ones reported in Figure \ref{ckb}. For the iterated bilateral filter, we use (b) to compute the range filter instead of using (a), however the averaging is performed on (a). Thus the run time of the iterated filter is roughly twice that of the proposed filter. Notice that (d) has some granular artefacts particularly along the edges, even though its PSNR is within a dB of that of  (c). In this regard, we note that the SSIM index of (c) is $0.82$, while that of (d) is $0.76$} 
\label{house}
\end{figure*}

We now present some denoising results to demonstrate the superior denoising performance of the proposed filter.
 In Figure \ref{ckb}, we compare the proposed filter with the standard and the oracle filter on a synthetic test image. 
 Notice that the improvement in PSNR over the standard filter is more than $10$ dB. 
 This does not come as a surprise since this particular test image has a lot of sharp intensity transitions. 
 While the bilateral filter is already known to work well for such images, what this result demonstrates is that we can further improve its performance using the proposed modification. Moreover, note the PSNR of the proposed filter is close to that of the oracle bilateral filter (which uses the clean image to compute the range filter). 
 In Figure \ref{house}, we compare the proposed filter with the iterated bilateral filter. Notice that while the PSNR from the iterated filter is within a dB of that obtained using the proposed filter, the denoised image from the latter looks visibly better. This is also confirmed by the SSIM indices reported in the figure. 

\begin{figure*}[!htb]
\centering
\subfloat[\textit{Lena} ($\sigma=30, 18.57$ dB).]{\includegraphics[width=0.33\linewidth]{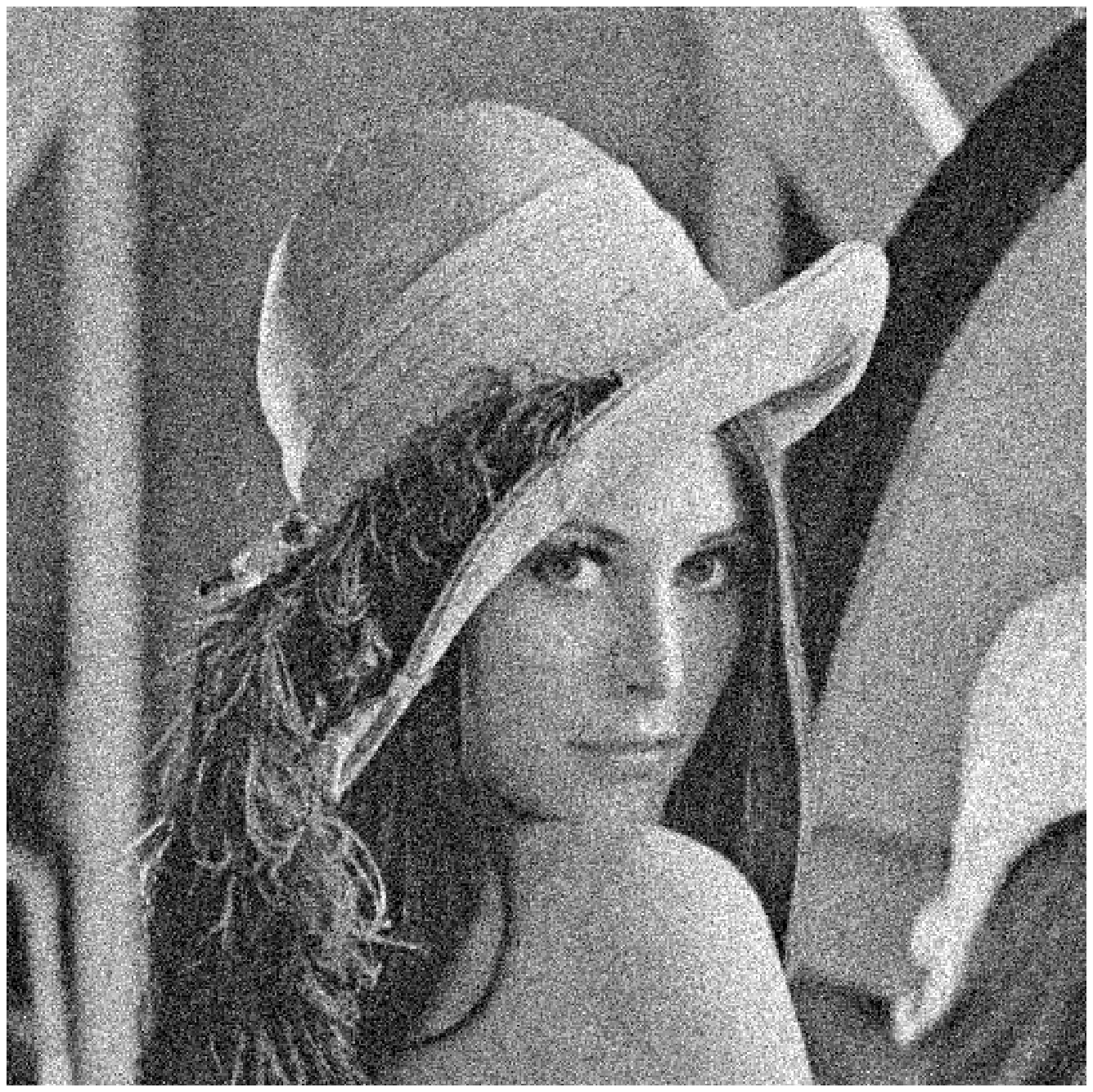}}  
\subfloat[SBF ($24.72$ dB, $\sigma_s= 2, \sigma_r = 40$).]{\includegraphics[width=0.33\linewidth]{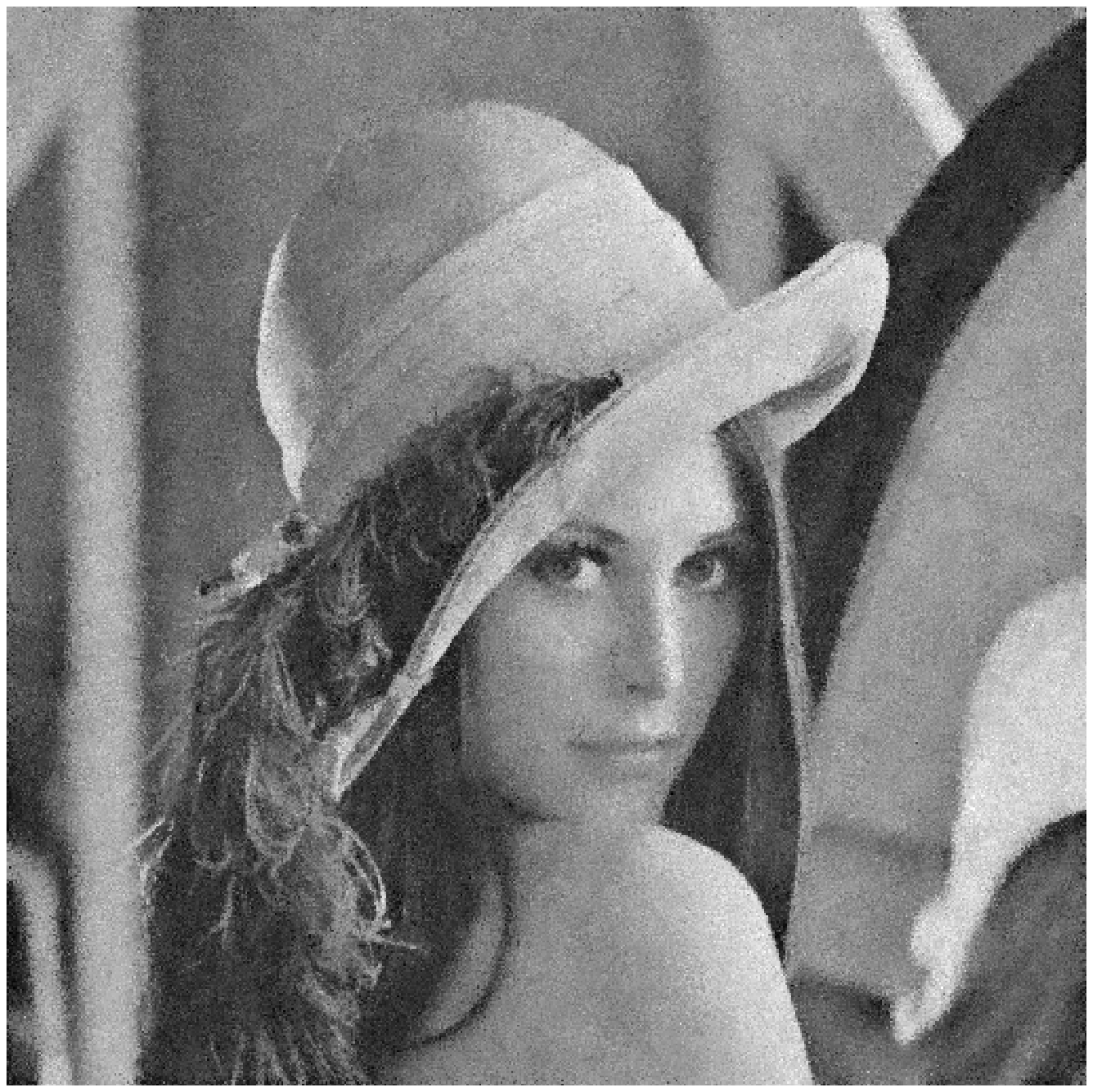}}   
\subfloat[IBF ($\bf{29.82}$ dB, $\sigma_s= 2, \sigma_r = 20$).]{\includegraphics[width=0.33\linewidth]{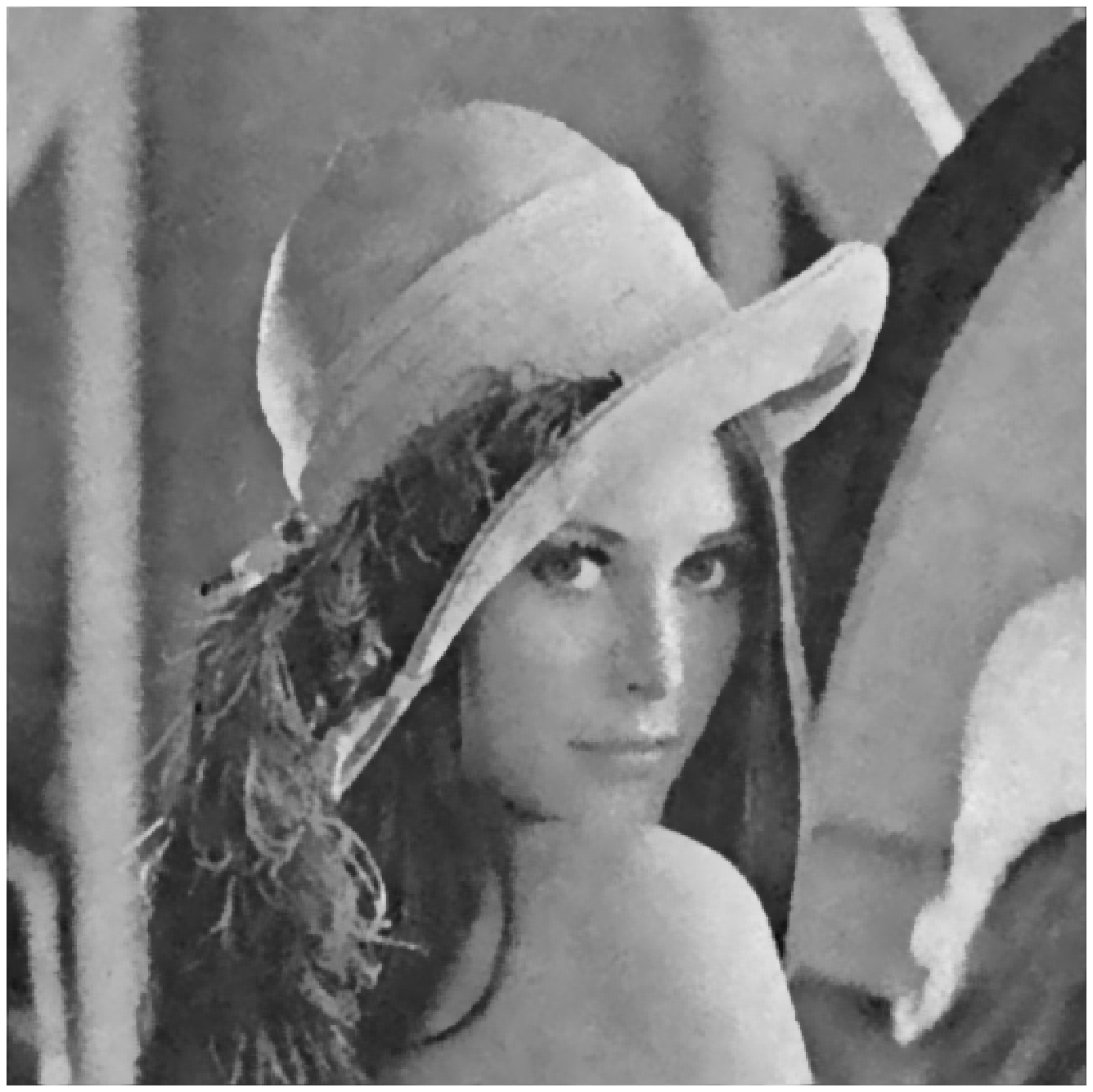}}  
\caption{Denoising results using standard bilateral filter (SBF) and improved bilateral filter (IBF). We tuned the parameters of SBF and IBF to get the optimal PSNR in either case. The parameters settings are indicated in the respective captions.} 
\label{visualComp1}
\end{figure*}

\begin{figure*}[!htb]
\centering
\subfloat[\textit{Peppers} ($\sigma=30, 18.55$ dB).]{\includegraphics[width=0.33\linewidth]{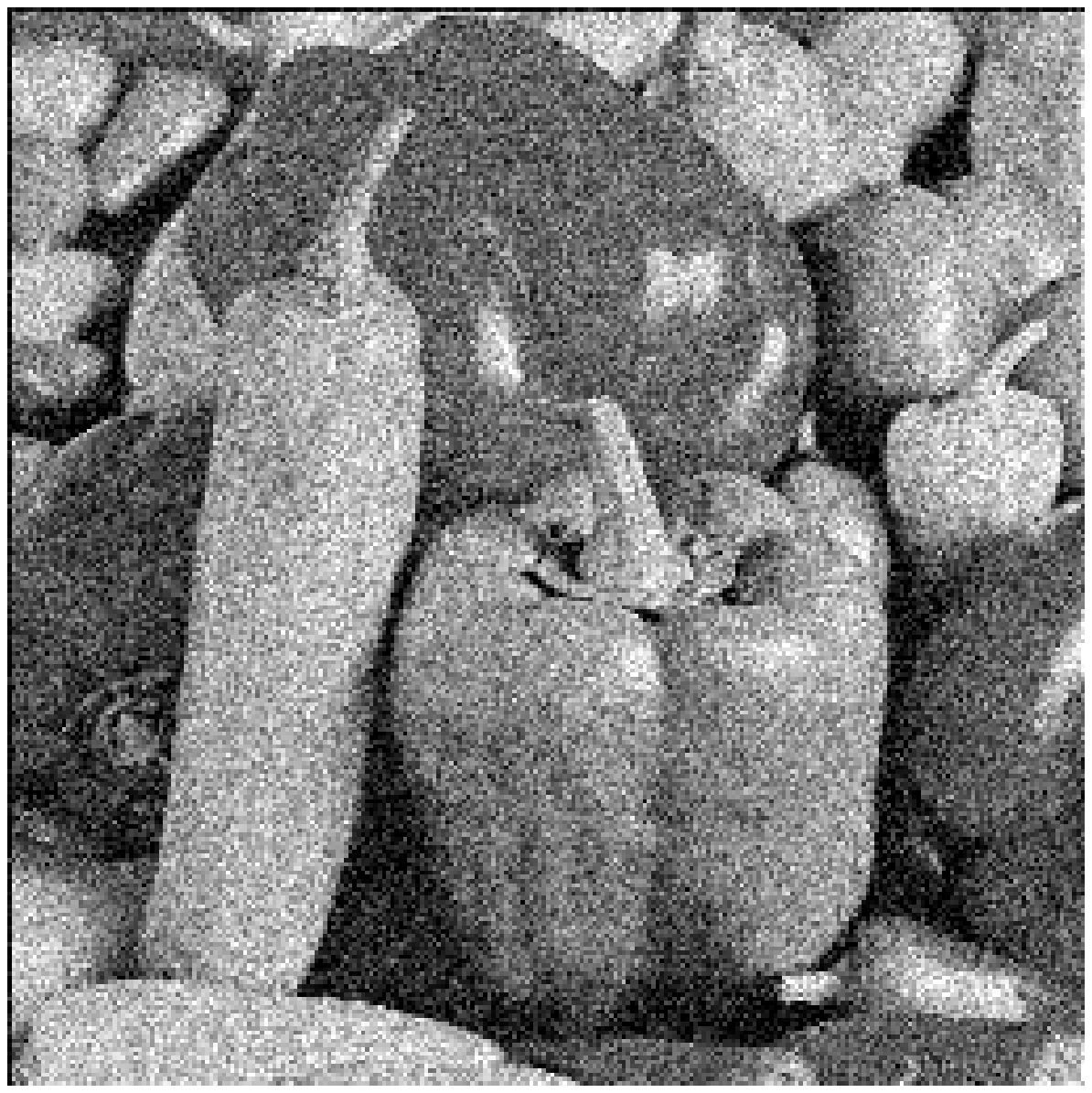}}  
\subfloat[SBF ($24.18$ dB, $\sigma_s= 2, \sigma_r = 40$).]{\includegraphics[width=0.33\linewidth]{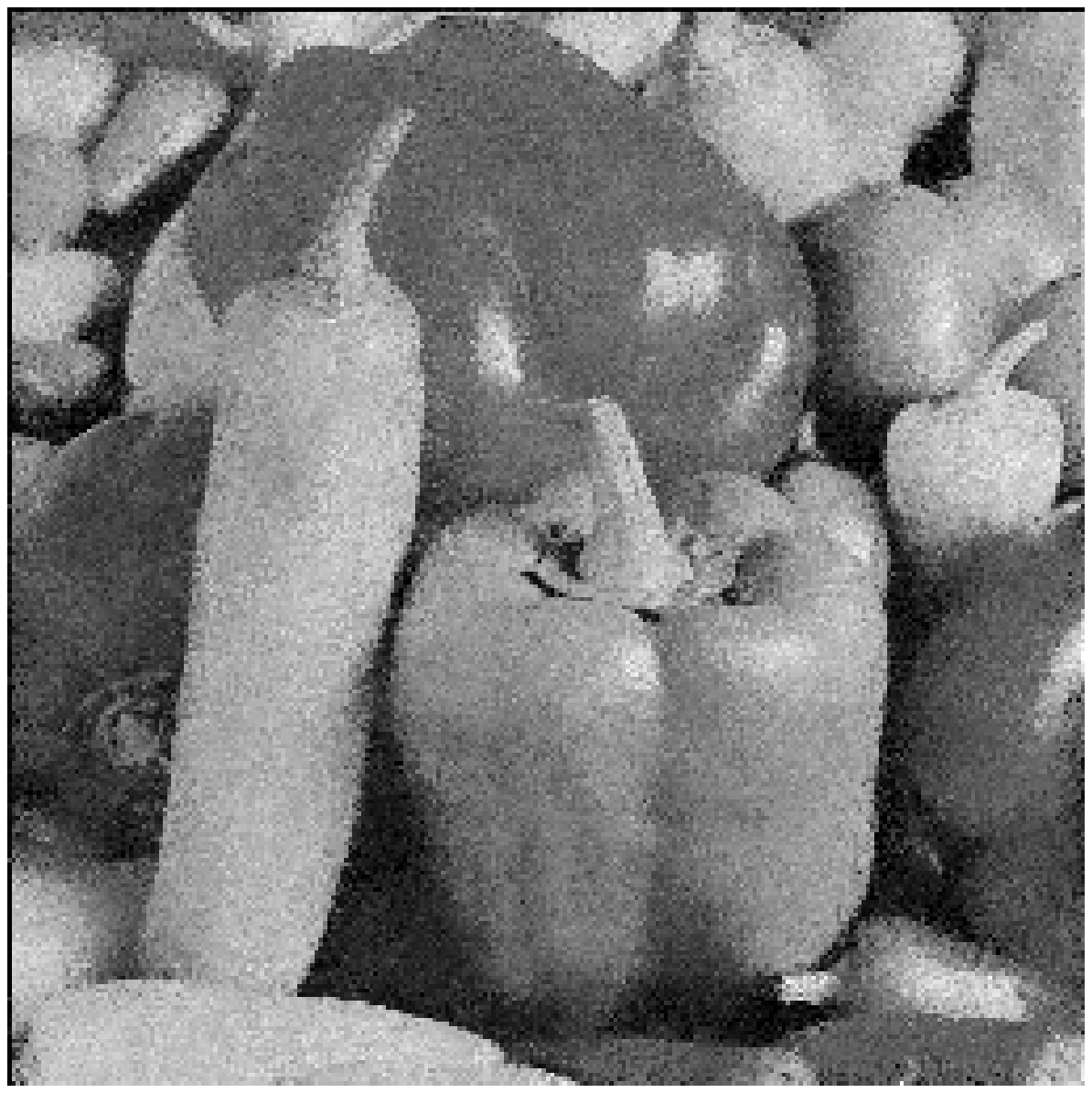}}   
\subfloat[IBF ($\bf{27.91}$ dB, $\sigma_s= 2, \sigma_r = 20$).]{\includegraphics[width=0.33\linewidth]{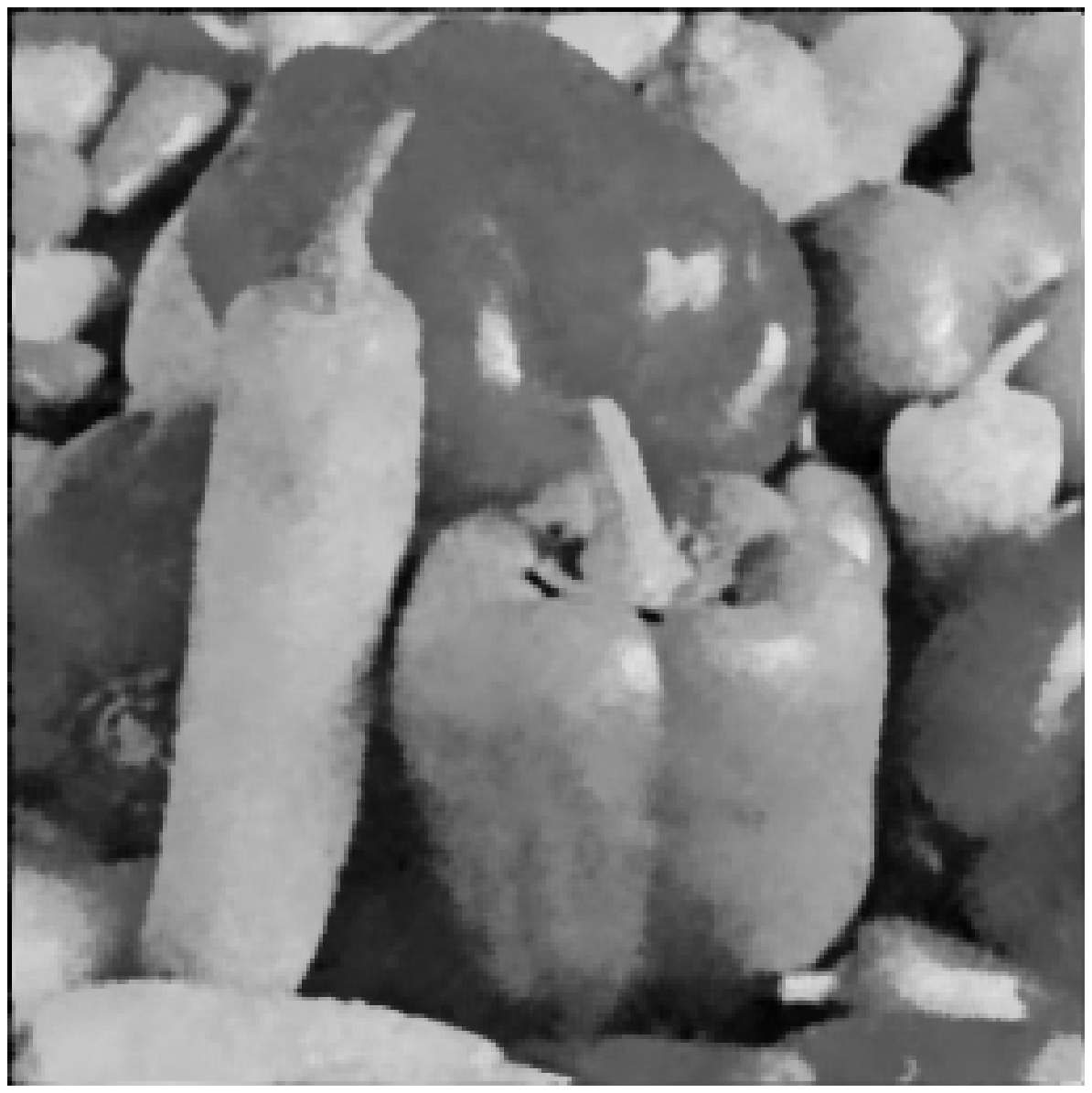}}  
\caption{Denoising results using standard bilateral filter (SBF) and improved bilateral filter (IBF). We tuned the parameters of SBF and IBF to get the optimal PSNR in either case. The parameters settings are indicated in the respective captions.} 
\label{visualComp2}
\end{figure*}

\begin{figure*}[!htb]
\centering
\subfloat[\textit{Cameraman} ($\sigma=35, 17.24$ dB).]{\includegraphics[width=0.33\linewidth]{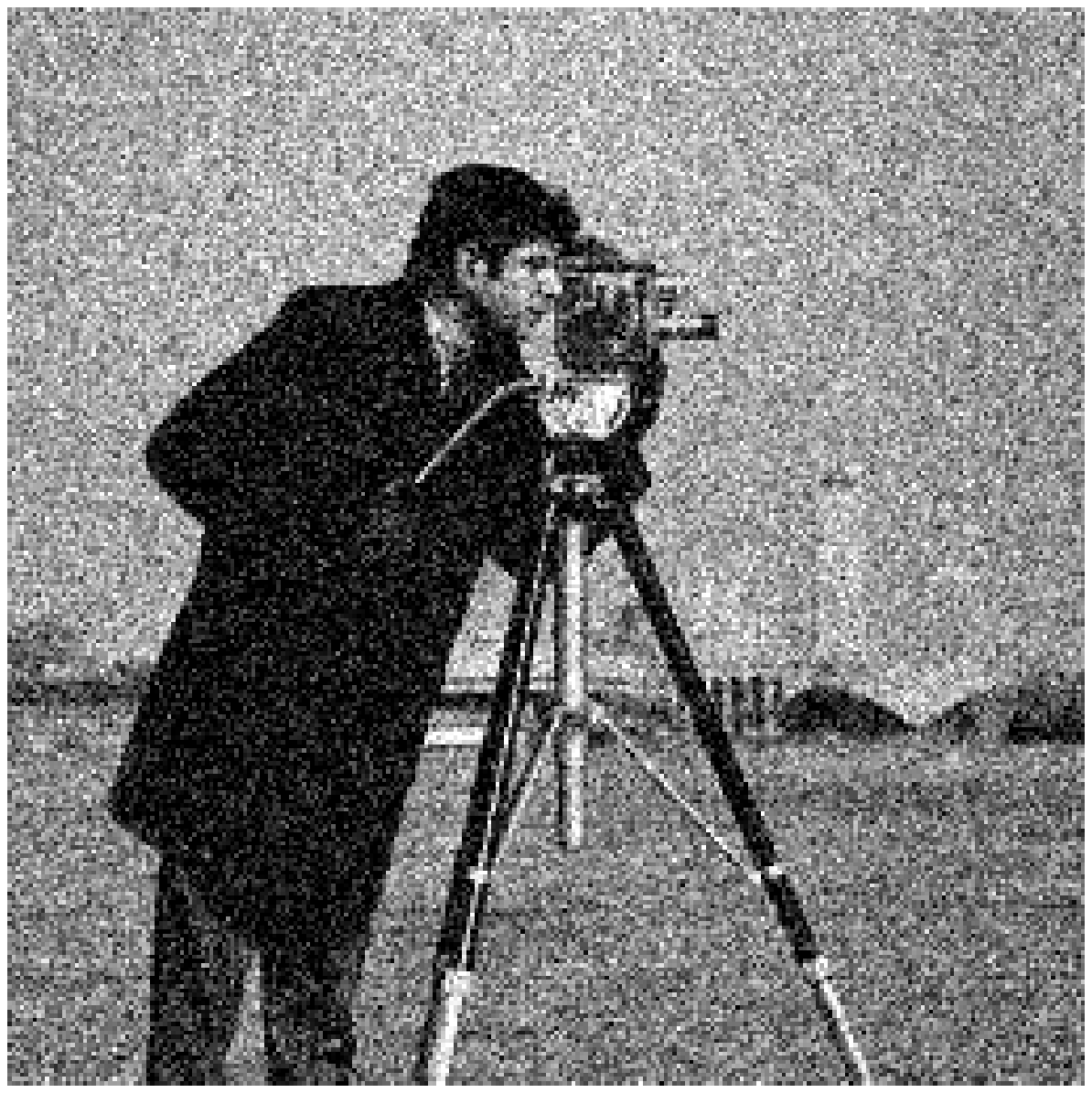}}  
\subfloat[SBF ($22.30$ dB, $\sigma_s= 2, \sigma_r = 40$).]{\includegraphics[width=0.33\linewidth]{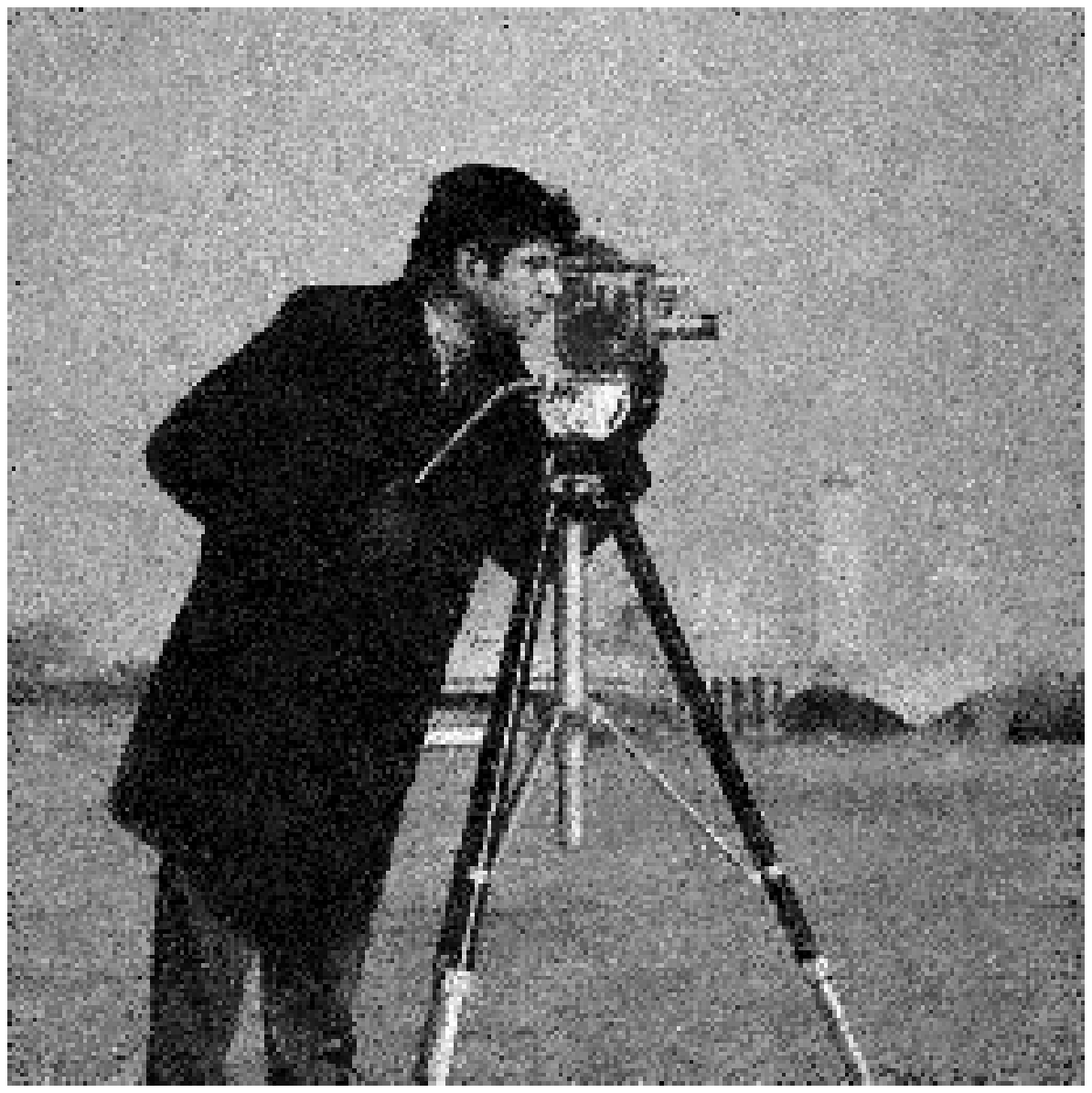}}   
\subfloat[IBF ($\bf{25.45}$ dB, $\sigma_s= 2, \sigma_r = 20$).]{\includegraphics[width=0.33\linewidth]{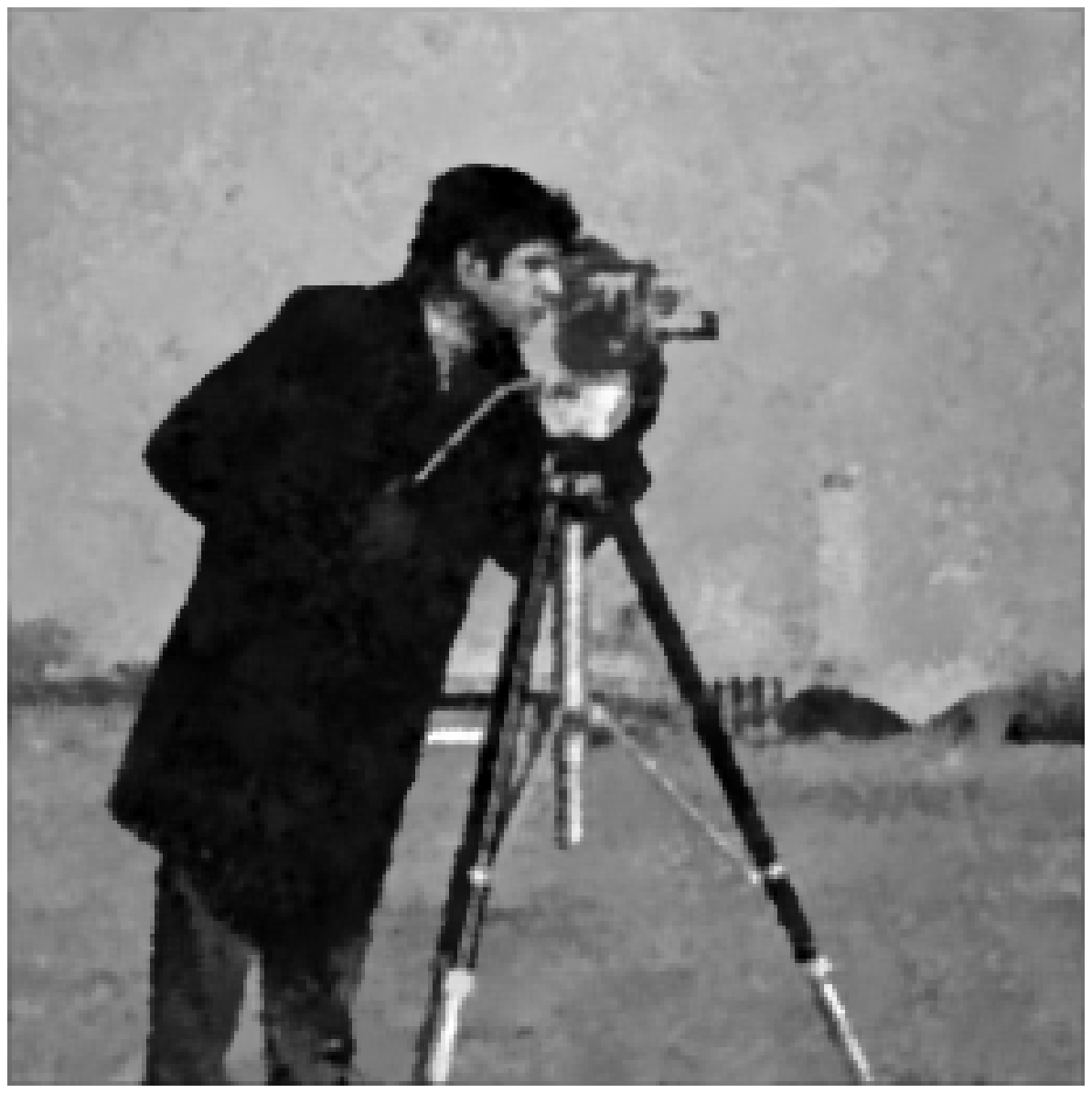}}
\caption{Denoising results using standard bilateral filter (SBF) and improved bilateral filter (IBF). We tuned the parameters of SBF and IBF to get the optimal PSNR in either case. The parameters settings are indicated in the respective captions.} 
\label{visualComp3}
\end{figure*}

 We next compare the denoising performance of the proposed modification with the standard bilateral filter on certain standard test images \cite{USCdatabase}. 
 The PSNR and SSIM indices of the proposed modification and the standard bilateral filter are reported in Table \ref{table2}.
 We independently optimize the standard and the improved bilateral filter with respect to $\sigma_s$ and $\sigma_r$.
Also, we use $L=1$ for the improved bilateral filter.
 Notice that the proposed filter starts to perform better beyond a certain noise level ($\sigma \approx 20$). 
 On the other hand, the SSIM improvement is already evident for all the images beyond $\sigma=15$. This is because, at low noise levels, the proposed box-filtering does more blurring than noise suppression, which brings down the overall signal-to-noise ratio. Indeed, when the noise floor is small, the corrupted image is already a good proxy for the clean image. However, notice that the improvement in SSIM is quite significant for all the images at large noise levels, and the PSNR improvement is often as large as $5$ dB. 
For a visual comparison, some of the results of the denoising experiments are shown in Figures \ref{visualComp1},  \ref{visualComp2}, and  \ref{visualComp3}.
In Table  \ref{table2}, we compare the proposed filter with some sophisticated image denoising methods cited in the introduction, namely, NLM \cite{Buades2005}, K-SVD \cite{KSVD}, and BM3D \cite{BM3D}. NLM is essentially a patch-based extension of the bilateral filter, in which image patches (groups of neighbouring pixels) are used for comparing neighbouring pixels. The latter methods are based on sparse-coding and collaborative filtering and are significantly more sophisticated. 
What we find quite interesting is that, beyond a certain noise level, the proposed filter is competitive with NLM for most of the test images in Table \ref{table2}, except for \textit{Barbara} and \textit{Cameraman}. The denoising performance is in general inferior to K-SVD and BM3D. We do not find this surprising since (among other things) they heavily rely on the sparsity of natural images (in appropriate bases) to improve the PSNR by few extra dBs.
However, we believe that the present work is relevant in the context of the recent results in \cite{Knaus2014} and \cite{Morel2014}, where the bilateral filter is used to obtain state-of-the-art denoising results.

\section{Conclusion}
\label{sec:conclusion}

We demonstrated that by using a box-filtered image in the range filter, one can substantially improve the denoising performance of the bilateral filter, at almost no additional cost. 
While the basic idea is quite simple, it is nevertheless quite effective in improving the denoising performance of the filter. Exhaustive denoising results on test images were provided in this direction. This  address a well-known pathology of the bilateral filter, namely, that its denoising performance begins to degrade quickly with the increase in noise level. 
An interesting finding was that the proposed filter is often competitive with the computation-intensive non-local means filter.  
We also presented a fast algorithm for the proposed  filter that can dramatically cut down the run time. 
As future work, we plan to investigate how one can combine the standard and the proposed filter so as to consistently obtain the best denoising performance at all noise levels.

\begin{table*}
\setlength{\tabcolsep}{4pt}
\caption{Comparison of the Standard Bilateral Filter (SBF) and the Improved Bilateral Filter (IBF) in terms of PSNR and SSIM at noise levels $\sigma = 10, 15, 20, 25, \ldots,55,60$. 
We also compare the PSNR with those obtained using more sophisticated methods such as Non-Local Means (NLM) \cite{Buades2005}, K-SVD \cite{KSVD}, and BM3D \cite{BM3D}. 
For a fixed image and noise level, we tune $\sigma_s$ and $\sigma_r$  to independently optimize the PSNRs obtained using SBF and IBF. We also tune the parameters of NLM to get the optimal PSNR at each noise level. If the PSNR obtained using IBF is higher than that obtained using NLM, we mark the former in boldface.}  
\vspace{2mm}
\centering

\begin{tabular}{l  c rrrrrrrrrrr}  
\hline 
\hline
Image & Filter &\multicolumn{11}{c}{PSNR (dB)} \\

\hline
\hline  

&SBF    &31.41  &28.78  &27.13  &25.51  &23.63  &21.93 &20.30   &18.76 &17.40 &16.21 &15.10 \\ 
&IBF     &25.60 &25.45  &25.18  &24.85   &24.47 &24.15 &23.89 &23.60 &23.42 &23.24  &22.96  \\  
\textit{Barbara}
&NLM    &33.04 &30.86  &29.27  &27.99   &27.16 &26.42 &25.74 &25.04 &24.50 &24.04  &23.67 \\  
&K-SVD  &34.42 &32.27  &30.76  &29.44   &28.40 &27.43 &26.61 &25.87 &25.23 &24.72  &24.18  \\  
&BM3D    &34.98 &33.05  &31.68  &30.55   &29.63 &28.76 &27.88 &27.63 &26.99 &26.54  &26.13  \\  

\hline
\hline

&SBF    &33.58  &31.60 &29.74  &27.30  &24.78  &22.70  &20.80  &19.15  &17.66   &16.36   &15.25 \\  
&IBF   &33.27  &\textbf{32.49} &\textbf{31.49}  &\textbf{30.59}  &\textbf{29.79}  &\textbf{29.19}  &\textbf{28.66}  &\textbf{28.05}  &\textbf{27.62}  &\textbf{27.17}  &\textbf{26.62} \\  
\textit{Lena}
&NLM    &34.06 &32.33  &30.99  &29.89   &29.07 &28.39 &27.74 &27.18 &26.72 &26.23  &25.85 \\  
&K-SVD  &35.50 &33.65  &32.40  &31.27   &30.44 &29.67 &29.01 &28.38 &27.79 &27.31  &26.90  \\  
&BM3D    &35.88 &34.20  &33.02 &32.03  &31.16   &30.47 &29.76 &29.44 &28.96 &28.57 &28.16   \\  

\hline
\hline

&SBF    &33.73 &33.70   &29.64  &27.18  &24.67  &22.62  &20.74  &19.06  &17.61  &16.36 &15.22 \\ 
&IBF    &33.17  &32.39  &31.38  &30.50  &\textbf{29.79}   &\textbf{29.11}   &\textbf{28.49}   &\textbf{27.72}   &\textbf{27.26}   &\textbf{26.82}  &\textbf{26.28}  \\ 
\textit{House}
&NLM    &34.63 &33.00  &31.63  &30.56   &29.34 &28.47 &27.69 &26.93 &26.36 &25.70  &25.00 \\  
&K-SVD    &35.89 &34.39  &33.07  &32.15   &31.29 &30.36 &29.59 &28.89 &27.90 &27.28  &27.08  \\ 
&BM3D  &36.80 &35.05  &33.71  &32.86   &32.15 &31.30 &30.88 &30.16 &29.99 &29.12  &28.69  \\

\hline
\hline

&SBF    &32.97  &30.74  &28.80  &26.54  &24.14  &22.01  &20.20  &18.64   &17.32  &16.08  &15.00  \\ 
&IBF    &31.31  &30.61  &\textbf{29.80}   &\textbf{28.72}   &\textbf{27.93}   &\textbf{27.03}  &\textbf{26.34}  &\textbf{25.69}  &\textbf{25.14}   &\textbf{24.65}  &\textbf{24.22}  \\ 
\textit{Peppers}
&NLM    &32.91 &30.71  &29.23  &28.06   &27.14 &26.26 &25.51 &24.98 &24.40 &23.82  &23.35 \\  
&K-SVD  &34.27 &32.34  &30.87  &29.73   &28.87 &28.10 &27.33 &26.79 &26.13 &25.66  &25.00  \\  
&BM3D    &34.70 &32.68  &31.27  &30.21   &29.21 &28.54 &27.67 &27.23 &26.75 &26.26  &25.89  \\  

\hline 
\hline

&SBF    &32.03  &29.90   &28.40   &26.39   &24.21  &22.24   &20.52   &18.93   &17.50   &16.26   &15.17  \\ 
&IBF    &29.96   &29.55    &\textbf{28.90}   &\textbf{28.15}   &\textbf{27.46}   &\textbf{26.90}   &\textbf{26.45}   &\textbf{25.93}   &\textbf{25.56}   &\textbf{25.16}   &\textbf{24.81}  \\ 
\textit{Boat}
&NLM    &31.93 &29.93  &28.57  &27.60   &26.90 &26.25 &25.68 &25.12 &24.58 &24.19  &23.84 \\  
&K-SVD  &33.63 &31.69  &30.36  &29.24   &28.42 &27.67 &27.01 &26.46 &25.95 &25.49  &25.07  \\  
&BM3D    &33.88 &32.09  &30.79  &29.81   &29.06 &28.35 &27.69 &27.10 &26.73 &26.27  &25.98  \\

\hline
\hline

&SBF    &32.65  &30.25   &28.54   &26.33  &24.23   &22.24    &20.44    &18.91   &17.40   &16.19   &15.08 \\ 
&IBF    &27.58  &27.35  &26.98   &26.49   &25.91   &25.41   &24.96   &24.63  &24.27  &23.89   &\textbf{23.62} \\ 
\textit{Cameraman}
&NLM    &32.61 &30.00  &28.57  &27.70   &27.01 &26.26 &25.39 &24.75 &24.28 &23.92  &23.22 \\  
&K-SVD  &33.73 &31.37  &29.96  &28.87   &28.00 &27.33 &26.69 &26.30 &25.74 &25.16  &24.92  \\  
&BM3D    &34.16 &31.84  &30.41  &29.53   &28.60 &27.84 &27.07 &26.63 &25.98 &25.72  &25.38  \\  
\end{tabular}

\begin{tabular}{l  c rrrrrrrrrrr}  
\hline 
\hline

&  &\multicolumn{11}{c}{SSIM (\%)} \\

\hline
\hline
&SBF    &89.51  &82.24  &75.21  &64.01  &51.54  &42.2  &35.15  &29.51  &24.77  &21.29  &18.39 \\
\raisebox{1.5ex}{\textit{Barbara}} 
&IBF  &76.36  &75.16  &73.00  &69.87  &67.61 &65.44  &63.71  &61.97  &59.87  &59.30  &56.51\\

\hline 
\hline
&SBF    &88.45  &84.12  &74.56  &60.29   &45.20  &34.22  &26.54  &21.18  &16.69  &13.80  &11.55  \\
\raisebox{1.5ex}{\textit{Lena}} 
&IBF  &87.80  &86.27  &83.52  &81.68   &79.89    &78.51   &77.14  &75.38   &72.84  &71.49   &68.01 \\

\hline
 \hline
&SBF    &86.92   &83.50   &74.02    &59.27   &45.17   &34.45   &27.19   &21.78   &17.69   &15.04   &12.86 \\
\raisebox{1.5ex}{\textit{House}} 
&IBF  &86.45   &84.94   &82.37   &81.33   &80.12  &78.78   &76.89   &73.50  &72.19   &70.69  &67.83 \\

\hline
\hline
&SBF    &90.86   &86.06   &78.37   &66.02   &52.56   &42.49   &34.40  &27.26   &23.14   &19.77  &17.23  \\
\raisebox{1.5ex}{\textit{Peppers}} 
&IBF  &89.83  &88.44   &86.01   &82.09  &80.29  &75.76  &73.52  &71.98  &72.03  &70.29  &66.98 \\

\hline
\hline

&SBF    &85.62  &79.31  &73.39  &62.01  &50.20  &38.88  &31.67  &25.85  &21.42  &17.97  &15.42  \\
\raisebox{1.5ex}{\textit{Boat}} 
&IBF  &80.51  &79.47  &77.30  &74.20  &71.86  &69.30  &67.72  &65.61  &63.22 &61.70 &59.99 \\

\hline
\hline
&SBF    &88.91  &82.21  &75.15  &61.39  &49.36  &39.14  &32.06  &27.23  &23.05  &20.21  &17.59  \\ 
\raisebox{1.5ex}{\textit{Cameraman}} 
&IBF  &84.07  &82.83  &80.09 &76.18  &73.76  &73.83  &73.61  &68.85  &69.46  &64.46  &62.49 \\

\hline
\hline                          

\end{tabular}
\label{table2}
\end{table*}

\bibliographystyle{IEEEbib}

\end{document}